\documentclass[runningheads]{llncs}

\usepackage[mobile]{eccv}

\usepackage{eccvabbrv}

\usepackage{graphicx}
\usepackage{booktabs}

\usepackage[accsupp]{axessibility}  

\usepackage{hyperref}

\usepackage{orcidlink}

\newcommand{\even}{{\text{even}}}
\newcommand{\odd}{{\text{odd}}}

\usepackage{bibunits}

\begin{document}

\title{Equivariant Symmetry-Aware Head Pose Estimation for Fetal MRI} 


\author{Ramya Muthukrishnan\inst{1}\orcidlink{0000-0001-5740-5963} \and
Borjan Gagoski\inst{2}\orcidlink{0000-0003-3777-2621} \and
Aryn Lee\inst{2}\orcidlink{0009-0002-5793-224X} \and P. Ellen Grant\inst{2}\orcidlink{0000-0003-1005-4013} \and Elfar Adalsteinsson\inst{1}\orcidlink{0000-0002-7637-2914} \and Benjamin Billot\inst{3}\orcidlink{0000-0002-3018-1282} \and Polina Golland \inst{1}\orcidlink{0000-0003-2516-731X}}

\authorrunning{Ramya Muthukrishnan et al.}

\institute{MIT Computer Science and Artificial Intelligence Laboratory, Cambridge, USA \\
\email{\{ramyamut,elfar,polina\}@mit.edu}\and
Boston Children's Hospital and Harvard Medical School, Boston, USA \\
\email{\{borjan.gagoski,aryn.lee,ellen.grant\}@childrens.harvard.edu} \and
Inria, Université Côte d'Azur, Sophia Antipolis, France \\
\email{bbillot@inria.fr}}

\maketitle

\begin{abstract}
  We present E(3)-Pose, a novel fast pose estimation method that jointly and explicitly models rotation equivariance and object symmetry. Our work is motivated by the challenging problem of accounting for fetal head motion during a diagnostic MRI scan. We aim to enable automatic adaptive prescription of diagnostic 2D MRI slices with 6-DoF head pose estimation, supported by rapid low-resolution 3D MRI volumes acquired before each 2D slice. Existing pose estimation methods struggle to generalize to clinical volumes due to pose ambiguities induced by inherent anatomical symmetries, as well as low resolution, noise, and artifacts. In contrast, E(3)-Pose captures anatomical symmetries and rigid pose equivariance by construction, and yields robust estimates of the fetal head pose. Our experiments on publicly available and representative clinical fetal MRI datasets demonstrate the superior robustness and generalization of our method across domains. Crucially, E(3)-Pose achieves state-of-the-art accuracy on clinical MRI volumes, supporting future clinical translation. Our implementation is publicly available\footnote[1]{\textbf{Code:} \href{https://github.com/MedicalVisionGroup/E3-Pose}{https://github.com/MedicalVisionGroup/E3-Pose}\\\textbf{Project page:} \href{ramyamut.github.io/e3-pose}{ramyamut.github.io/e3-pose}}.
  \keywords{6-DoF pose estimation \and object symmetry \and fetal MRI}
\end{abstract}

\section{Introduction}
\label{sec:intro}

\begin{figure*}[!t]
\centerline{\includegraphics[width=\textwidth]{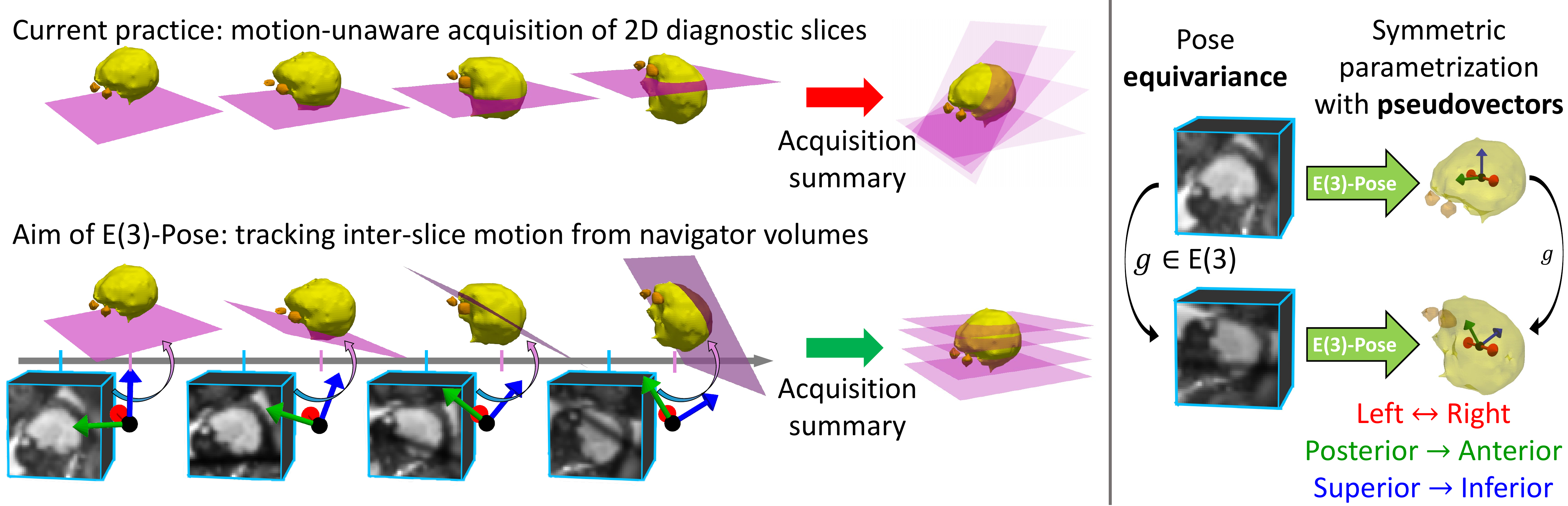}}
\caption{\textbf{E(3)-Pose is a rotation-equivariant and symmetry-aware framework for 6-DoF pose estimation.} \textit{Left}: a rapid navigator volume is inserted between every two 2D diagnostic MRI slices. Our aim is to estimate the fetal head pose to adjust imaging plane prescription in real time. \textit{Right}: To enable robust performance, the network architecture employs E(3)-equivariant convolutional filters to capture pose equivariance and pseudovectors to account for left-right head symmetry, respectively.}
\label{fig:teaser}
\end{figure*}

Rapid 6-DoF object pose estimation enables real-time navigation in robotic manipulation~\cite{billard2019}, autonomous driving~\cite{chiu2021}, and image-guided surgery~\cite{yaniv2006}. Here, we present a novel method for rapid object pose estimation from 3D images based on an architecture that captures pose equivariance and object symmetries by construction. We are motivated by the problem of real-time motion estimation in fetal MRI, where stacks of high-resolution 2D MRI slices are acquired to assess fetal development and detect pathology~\cite{glenn2009,gholipour2014} (Appendix~\ref{sec:interleaved}). Due to inter-slice fetal motion, oblique slices and spatial coverage gaps in the acquired stack present challenges for radiological assessment and often necessitate re-acquisition of the entire stack~\cite{malamateniou2013,xu2023,white2010}. One solution is to estimate the fetal head pose from rapid, low-resolution, low signal-to-noise ratio (SNR) 3D MRI ``navigator'' volumes inserted into the MRI sequence before every 2D slice. The aim is to use the pose estimated from each navigator volume to adaptively adjust the imaging plane of the next slice
(Fig.~\ref{fig:teaser})~\cite{hess2011,tisdall2012,gagoski2016}.
Estimating fetal head pose from navigator volumes is particularly challenging. In addition to low resolution and SNR, navigator volumes contain an imaging artifact, a dark plane in the position of the previous (high energy) slice that may obstruct relevant anatomy (e.g., eyes) that is crucial for pose disambiguation. Moreover, underdeveloped neuroanatomy coupled with low resolution creates an approximate left-right symmetry, introducing pose ambiguities~\cite{zhan2013}. Finally, the lack of available navigator data necessitates training on artifact-free, high-resolution, high-SNR research volumes, thus creating cross-domain generalization challenges. 

Several existing methods detect pre-specified anatomical landmarks to compute pose~\cite{hoffmann2021,silva2024}, but landmark visibility is not guaranteed in clinical data. Another class of methods formulates the problem as rigid registration of a moving volume to a template~\cite{moyer2021,billot2024,wang2025}. However, artifacts perturb the intensity distributions of the navigator volumes relative to the template, leading to suboptimal registration. While canonical pose regression with convolutional neural networks (CNNs) can improve robustness to reduced anatomical visibility and imaging artifacts~\cite{xiang2018,salehi2019,gao2020_2}, these methods are prone to overfitting even with data augmentations and are sensitive to pose ambiguities.

\subsubsection{Contributions.} We propose E(3)-Pose, a framework for accurate and generalizable head pose estimation from fetal brain volumes (Fig.~\ref{fig:teaser}). In contrast to standard convolutions that are only equivariant under translations, we leverage an E(3)-equivariant CNN (E(3)-CNN) for rotation regression to also account for equivariance under rotations~\cite{weiler2018, geiger2022}. Moreover, we introduce a novel equivariant rotation parametrization that uses pseudovectors~\cite{hauser1986} to model left-right head symmetry. E(3)-Pose achieves competitive performance on publicly available research-quality datasets, and state-of-the-art performance on \textit{clinically} representative datasets. Crucially, our experiments demonstrate that constraining the network architecture to explicitly model physical symmetries provides better generalization to out-of-distribution data and stabilizes pose estimates in ambiguous cases with low anatomical visibility. While this paper focuses on fetal MRI, our method holds promise for object localization tasks that involve symmetries, limited training data, and real-world applications with noisy sensors.

\section{Related Work}

\subsubsection{Landmark-based Pose Estimation.} Landmark registration involves algorithmic detection of predefined anatomical structures to estimate pose relative to a canonical anatomical frame~\cite{hill2001,hoffmann2021,silva2024}. To mitigate the dependence on visibility of specific anatomy, recent approaches train a neural network to learn an overcomplete set of landmarks~\cite{wang2023, billot2024}. Nevertheless, learned landmarks display sensitivity to perturbations of volume intensities, posing problems when fast acquisitions introduce disruptive image artifacts. In contrast, we directly regress pose, circumventing these challenges.

\subsubsection{Pose Regression.} Early pose regression approaches parametrized rotations using axis-angle vectors and quaternions, which suffer from discontinuities and double-cover, respectively~\cite{salehi2019, grassia1998}. It has been shown that high-dimensional parametrizations that are continuous functions over SO(3) offer smoother loss landscapes and robustness to noise~\cite{geist2024,zhou2019}. Existing methods that adopt this strategy regress two basis directions~\cite{faghihpirayesh2023,zhou2019} or a deformation field~\cite{gopinath2024} of the canonical object frame relative to the input volume. E(3)-Pose similarly regresses a continuous, overcomplete parametrization by using three basis directions.

\subsubsection{Equivariant Networks.} A function is equivariant under a group of transforms if it commutes with every transform in the group. In the context of CNNs, standard convolutions are already equivariant under translations. There has been significant progress toward convolutions that are equivariant under a broader group, including SE(3) and E(3)~\cite{weiler2018, geiger2022,deng2021, cohen2016,cohen2017,weiler2019, thomas2022,worrall2017,cohen2018}. Specifically, these architectures restrict the hypothesis space to strictly adhere to inherent symmetries, rather than learning equivariance through data augmentation. Theoretical analysis shows that equivariant architectures are more sample-efficient, reducing overfitting especially when training data is limited~\cite{lawrence2022, sannai2019, bulusu2022}. By leveraging an E(3)-equivariant CNN for pose estimation, E(3)-Pose robustly generalizes to noisy navigator data.

Existing equivariant methods for 6-DoF pose estimation either predict the relative transform between image pairs~\cite{moyer2021, wang2022,billot2024,lee2022,zhang2020} or formulate rotation estimation as classification over a discretized set of 3D rotations, limiting the precision of the produced pose estimate~\cite{li2021, musallam2022, howell2023,lee2024,chen2021}. Furthermore, these methods do not explicitly account for symmetric ambiguities in their network architectures. In contrast, E(3)-Pose regresses a continuous output over SO(3) and handles object symmetry in the construction of this parametrization.

\subsubsection{Object Symmetries.} Rotational and reflectional object symmetries can lead to pose ambiguities. One solution is to train a network to implicitly learn object symmetries via probabilistic outputs, and to heuristically select the most probable pose during inference~\cite{hodan2020, zhao2023}. Yet, these methods suffer from discontinuities in the network output between symmetrically equivalent poses~\cite{richterklug2021}. Although training with symmetry-invariant losses partially alleviates this problem~\cite{xiang2018}, these instabilities are more robustly eliminated by incorporating object symmetry into the pose parametrization itself. Namely, this strategy regresses an output that remains \textit{invariant} under the object symmetry group, while maintaining continuity over SO(3)~\cite{pitteri2019,saxena2007,richterklug2021}. To handle head pose ambiguities, E(3)-Pose models the basis direction along the axis of left-right symmetry as a pseudovector~\cite{hauser1986}, which maintains both invariance under left-right reflections and continuity over SO(3). While pseudovectors have been used to model rotation axes in pose estimation~\cite{lee2019}, our method is the first to use them to address object symmetries.

\subsubsection{Automated Slice Prescription.} Existing implementations use the pose estimated from a single volume acquired at the start of the sequence to prescribe all slices in the stack~\cite{hoffmann2021,silva2024}. While this approach accurately prescribes slices according to the initial head pose, it does not account for inter-slice motion. In contrast, E(3)-Pose aims to adjust the imaging plane for every slice in the stack, based on the head pose estimated from the preceding navigator volume.

\section{Preliminaries}

In this section, we introduce key theoretical concepts in the design of E(3)-CNNs to provide necessary background or E(3)-Pose.

\subsubsection{Tensor Fields and E(3) Representations.} Let $f\!:\!\mathbb{R}^3\!\rightarrow\!\mathbb{R}^{d}$ be a $d$-dimension\-al tensor field. 3D Euclidean group action $g\!=\!g_t\circ g_r$ comprises rotation/reflection $g_r\!\in\!\text{O(3)}$ and translation $g_t\!\in\!\mathbb{R}^3$. When the coordinate system is transformed by $g$, the tensor $f(x)$ at point $x\!\in\!\mathbb{R}^3$ is modified in two ways: it is moved to the location $g^{-1}x$ and the tensor is independently rotated/reflected by $g_r$~\cite{weiler2018,geiger2022}. Formally, operator $[\pi_f(g)f](x)\!=\!\rho_f(g_r)f(g^{-1}x)$ transforms field $f$ for any $g\!=\!g_t\circ g_r\!\in \!\text{E(3)}$. In this formulation, $\rho_f\!:\!\text{O(3)}\!\rightarrow\!\mathbb{R}^{d \times d}$ is a \emph{representation}; it is a matrix acting linearly on tensor $f(x)$ that describes how it transforms under $g_r$.

\subsubsection{Irreducibility.} A representation $\rho$ and the tensors it transforms are \emph{irreducible} if there is no change of basis $Q\!\in\!\mathbb{R}^{d\times d}$ such that $Q\rho(g_r)Q^{-1}$ is block-diagonal for all $g_r\!\in\! \text{O(3)}$. Irreducible representations and tensors are characterized by order $l$ and parity, defining how $f(x)$ transforms under proper rotations and inversion (i.e., reflection across all three planes) of the coordinate system, respectively~\cite{geiger2022}. Even parity indicates no change in $f(x)$ under inversion, while odd parity indicates a sign change in $f(x)$. While one can define tensors and representations of any order and parity, here we focus on three types: (\textit{i}) scalars $f(x)\!\in\!\mathbb{R}$ (order $l\!=\!0$, even parity) have representation $\rho^{l=0}_{\even}(g_r)\!=\!1$; (\textit{ii}) vectors $f(x)\!\in\!\mathbb{R}^3$ ($l\!=\!1$, odd parity) have representation $\rho^{l=1}_{\odd}(g_r)\!=\!M(g_r)$, where $M\!:\! \text{O(3)}\! \rightarrow \!\mathbb{R}^{3\times 3}$ is the matrix representation of an orthogonal transformation; and (\textit{iii}) pseudovectors $f(x)\!\in\!\mathbb{R}^3$ ($l\!=\!1$, even parity) have representation $\rho^{l=1}_{\even}(g_r)\!=\!(\det M(g_r))M(g_r)$ with a positive determinant to account for invariance under inversion~\cite{hauser1986}.

\subsubsection{Equivariance.} Let $F$ be a mapping between two tensor fields $f$ and $h$ associated with operators $\pi_f$ and $\pi_h$ and representations $\rho_{f}$ and $\rho_{h}$, respectively. Let $\tilde{F}$ be a mapping from $f$ to a single tensor $h(x)$ with representation $\rho_{h}$. $F$ and $\tilde{F}$ are equivariant under E(3) if and only if
\begin{equation}
\begin{aligned}
    F(\pi_{f}(g)f) &= \pi_{h}(g)F(f), \text{   for all } g \in \text{E(3)}.\\
    \tilde{F}(\pi_{f}(g)f) &= \rho_{h}(g_r)\tilde{F}(f), \text{   for all } g\!=\!g_t\circ g_r \in \text{E(3)}.
\end{aligned}
\label{eq:equivariance}
\end{equation}

\subsubsection{Equivariant Networks.} E(3)-CNNs can be built using successive equivariant layers, since their composition remains equivariant. Here we describe equivariant convolutional layers and defer pooling and non-linearities to Appendix~\ref{sec:e3cnn_arch}. For enhanced computation and time efficiency, state-of-the-art implementations of such CNNs operate on irreducible fields~\cite{weiler2018,geiger2022}. Let $\kappa_{l_fp_f,l_hp_h}(x)\in\mathbb{R}^{(2l_h+1)\times(2l_f+1)}$ be a 3D convolution kernel transforming field $f$ (order $l_f$, parity $p_f$) to field $h$ (order $l_h$, parity $p_h$). Then, $\kappa$ must satisfy Eq.~(\ref{eq:equivariance}) to maintain equivariance, resulting in the solution
\begin{equation}
\text{vec}(\kappa_{l_fp_f,l_hp_h}(x)) \!=\!\sum_{m \in M}\sum_{j \in J} w_{jm}\varphi_m(||x||)Q_jY_j(x/||x||),
\end{equation}
where vec indicates vectorization, $w_{jm}$ are coefficients, $\{\varphi_m\}$ are radially symmetric basis functions, $Q_j\!\in\!\mathbb{R}^{(2l_h+1)(2l_f+1)\times(2j+1)}$ are changes of the basis, $Y_j\!:\!S^2\rightarrow \mathbb{R}^{2j+1}$ are the degree-$j$ spherical harmonics, and $J\!=\!\{j\!:\!|l_f-l_h|\!\leq\!j\!\leq\!|l_f+l_h|, j \text{ is even}\}$ if $p_f\!=\!p_h$ or $J\!=\!\{j\!:\!|l_f-l_h|\!\leq\!j\!\leq\!|l_f+l_h|, j \text{ is odd}\}$ if $p_f\!\neq\!p_h$~\cite{weiler2018, geiger2022}. E(3)-CNNs are trained by learning the coefficients $w_{jm}$ rather than the kernel values themselves.

\begin{figure*}[!t]
\centerline{\includegraphics[width=\textwidth]{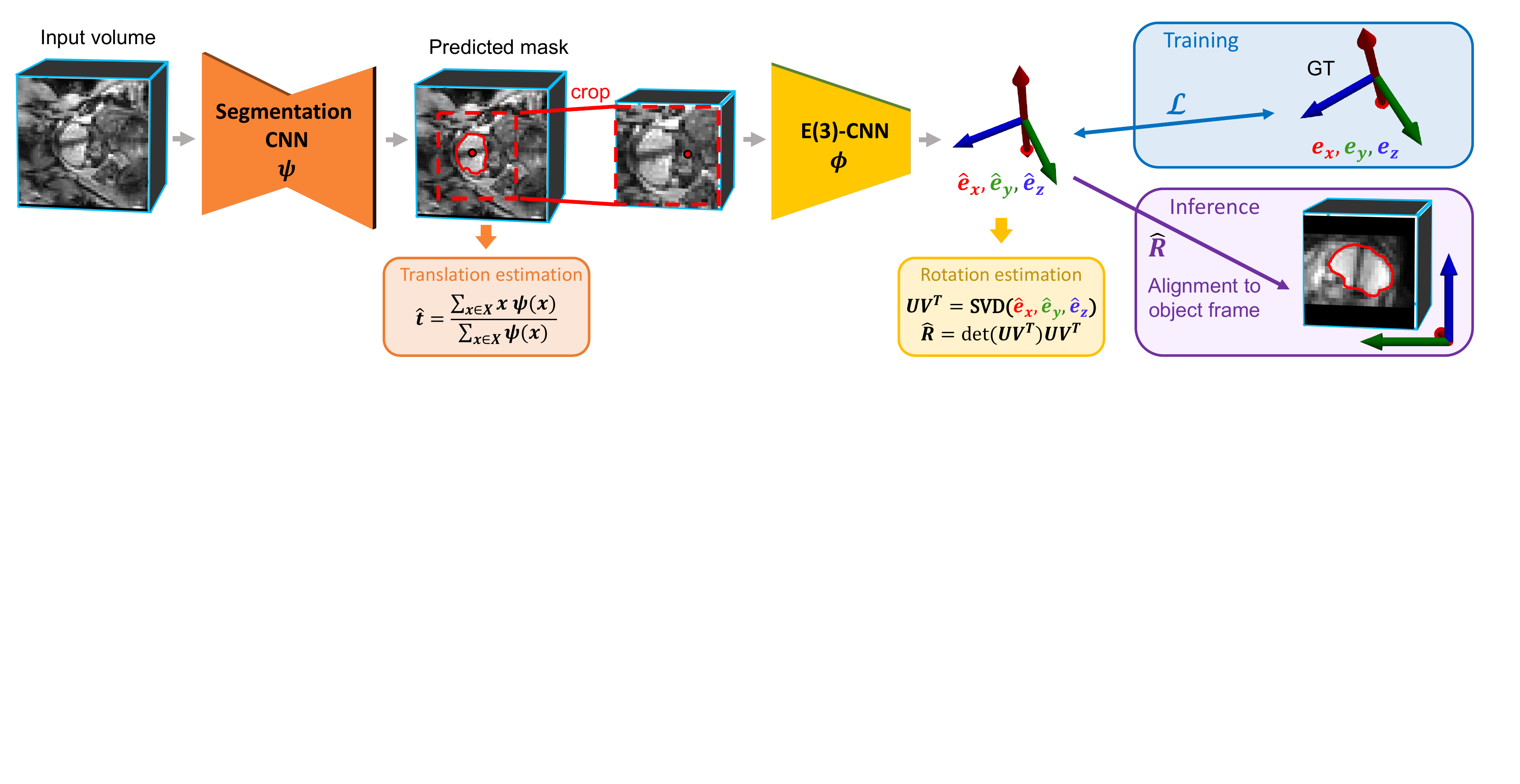}}
\caption{\textbf{Overview of E(3)-Pose.} We first train a CNN $\psi$ to segment the object. We estimate translation based on the center-of-mass of the predicted mask and then crop the 3D volumes around this mask using a 40\% margin. The cropped volumes are provided to an E(3)-CNN $\phi$ trained independently to regress the orthonormal basis of the object frame, parametrized as one pseudovector (red) and two vectors (blue and green). The output is later constrained to represent a rotation matrix by applying SVD and choosing the pseudovector $\hat{e}_x$ direction that results in a proper rotation without reflection (i.e., $\det(M(\hat{R}))=1$).}
\label{fig:overview}
\end{figure*}

\section{Methods}

Our goal is to estimate a 6-DoF canonical object pose $T$ from a 3D volume, with respect to the volume coordinate frame. We estimate the rigid pose $T\!=\!t\circ R$ by separating it into its rotation $R$ and translation $t$ components (Fig.~\ref{fig:overview}).

\subsection{Translation Estimation}

In rigid pose estimation, it is customary to place the origin of the object coordinate frame at the object center~\cite{xiang2018}. The translation $t$ can be computed from a binary object occupancy function $\psi: \mathbb{R}^3 \rightarrow \{0,1\}$: $t=\frac{\sum_{x\in X}x \psi(x)}{\sum_{x\in X}\psi(x)}$, where $X$ is the set of sampled points (i.e., voxel coordinates). Appendix~\ref{sec:unet_arch} describes our implementation of a CNN for fetal brain segmentation from whole-uterus MRI volumes that constructs the occupancy function $\psi$.
%
%

\subsection{Rotation Estimation}
\label{sec:rot}

\subsubsection{Function Equivariance.} We define $f\!:\!\mathbb{R}^3\!\rightarrow \!\mathbb{R}$ as the scalar field representing the input volume, with transformation operator $[\pi_{f}(g)f](x)\!=\!f(g^{-1}x)$; $h(R)\in\!\mathbb{R}^d$ as some $d$-dimensional tensor parametrization of the output rotation~$R$ with representation $\rho_h$; and $\phi$ as the rotation estimation function. Rotating/reflecting the input volume rotates/reflects its pose equivalently, requiring function equivariance detailed in Eq.~(\ref{eq:equivariance}):
\begin{equation}
\phi(f\circ g^{-1})\!=\!\rho_{h}(g_r)\phi(f), \:\:\: \text{for all} \:\: g\!=\!g_t\circ g_r \in \text{E(3).}
\end{equation}
Although in practice we require equivariance under SE(3) rather than the full group E(3), we show below that using an E(3)-CNN to implement $\phi$ enables us to jointly formulate $h(R)$ as equivariant under SE(3) and invariant under left-right reflections. Since E(3)-CNNs must operate on irreducible tensors, it suffices to parametrize $h(R)$ with irreducible tensors to guarantee the equivariance of $\phi$. 

%
%

\subsubsection{Rotation Parametrization.} Let $G_{\text{symm}}\!=\!\{g_s\in \text{O(3)}\!:\!f\circ g_s^{-1}\!=\!f\}$ be the object symmetry group~\cite{saxena2007}. To handle object symmetries, we aim to parametrize $h(R)$ such that $\rho_{h}(g_s)h(R)\!=\!h(R)$ for all $g_s\!\in\!G_{\text{symm}}$ and $R\!\in\!\text{SO(3)}$. Moreover, $h$ should remain continuous over SO(3). Here, we consider the case $G_{\text{symm}}\!=\!\{g_{\leftrightarrow}\}$, where $g_{\leftrightarrow}$ denotes left-right reflection about the left-right axis $e_{\leftrightarrow}$.

Formally, we define $h(R)$ as three orthonormal basis directions $\{e_x$, $e_y$,$e_z\}$ of the object frame ($d\!=\!9$), a continuous parametrization over SO(3):
\begin{equation}
\begin{aligned}
h(R)&=e_x\oplus e_y\oplus e_z, \text{  s.t. }  e_x\!=\!e_{\leftrightarrow} \text{  and  }  e_y,e_z\!\perp\!e_{\leftrightarrow},\\
\rho_{h}(g_r)\!&=\!(\det{M(g_r)})M(g_r)\oplus M(g_r) \oplus M(g_r)\\
&=\rho^{l=1}_{\even}(g_r)\oplus \rho^{l=1}_{\odd}(g_r) \oplus \rho^{l=1}_{\odd}(g_r),
\end{aligned}
\label{eq:rot_param}
\end{equation}
where $\oplus$ denotes columnwise and block-diagonal concatenation for vectors and matrices, respectively. Specifically, we decompose $h(R)$ into three irreducible tensors: a \textit{pseudovector} parametrizes the left-right direction, and two vectors parametrize two directions orthogonal to the left-right axis.
\begin{theorem}
$h(R)$ is invariant under $G_{\text{symm}}$.
\end{theorem}
\begin{proof}\leavevmode\vspace{-\baselineskip}
\begin{align*}
\rho_h(g_{\leftrightarrow})h(R)&= (\det M(g_{\leftrightarrow}))M(g_{\leftrightarrow})e_x \oplus M(g_{\leftrightarrow})e_y \oplus M(g_{\leftrightarrow})e_z\\
&= (\det M(g_{\leftrightarrow}))(-e_x) \oplus e_y \oplus e_z
= e_x \oplus e_y \oplus e_z = h(R),
\end{align*}
\end{proof}
where the second equality follows from the fact that following reflection, vectors orthogonal to the reflection axis are unchanged, and those that are parallel become inverted. Since we use an E(3)-CNN to estimate $h(R)$, $h(R)$ is guaranteed to be equivariant under $\text{SE(3)}\!\subset\!\text{E(3)}$.

\subsubsection{E(3)-CNN Regressor.} To handle both rotational equivariance and reflectional symmetry, we implement $\phi$ as an E(3)-CNN regressor that takes a scalar field as input and returns an output tensor that comprises one pseudovector $\hat{e}_x$ and two vectors $\hat{e}_y, \hat{e}_z$, normalized to unit length. Appendix~\ref{sec:e3cnn_arch} provides a detailed description of the network architecture. During training, we minimize the objective
\begin{equation} 
\begin{aligned}
\mathcal{L}((e_x,e_y,e_z),(\hat{e}_x,\hat{e}_y,\hat{e}_z)) \!&=\!\beta_x|\sin{\theta_x}|\!+\beta_y| \sin{\frac{\theta_y}{2}}|+\beta_z| \sin{\frac{\theta_z}{2}}|,
\end{aligned}
\label{eq:loss}
\end{equation}
where $(e_x,e_y,e_z)$ is the ground-truth (GT) orthonormal basis, $\theta_k\!=\!\arccos(\hat{e}_k\cdot e_k)$ for $k\in \{x,y,z\}$, and $\beta_x,\beta_y,\beta_z$ are weight hyperparameters. Our loss function accounts for the inversion symmetry of the pseudovector output $\hat{e}_x$ (Appendix~\ref{sec:e3cnn_loss}). During inference, we obtain an orthonormal output matrix by applying SVD to the predicted basis~\cite{levinson2020}. We then choose between $\hat{e}_x$ and $-\hat{e}_x$ in order to guarantee a positive determinant matrix, i.e., a proper rotation. This step is similar to prior work that heuristically selects the output pose from a set of symmetrically plausible poses~\cite{pitteri2019, hodan2020, zhao2023}. Here, the task involves choosing between right- and left-handed rotations.

\subsection{Implementation Details}

\subsubsection{Object Frame.} We place the origin of the target object frame at the center-of-mass (CoM) of the brain. We define the GT orthonormal basis $e_x, e_y$, and $e_z$ as the unit vectors pointing along the left$\rightarrow$right (L$\rightarrow$R), posterior$\rightarrow$anterior (P$\rightarrow$A), and inferior$\rightarrow$superior (I$\rightarrow$S) anatomical orientations of the brain, respectively (Fig.~\ref{fig:overview}). This follows Eq.~(\ref{eq:rot_param}), where $e_x$ is parallel and $e_y,e_z$ are orthogonal to the left-right axis, respectively. Appendix~\ref{sec:annotations} provides further details.

\subsubsection{Data Augmentation.} We augment training volumes with commonly used spatial (rigid transformations, scaling, flips) and intensity (bias field, gamma, Gaussian noise, low resolution) transforms in volumetric brain MRI~\cite{perezgarcia2021, billot2023}. To enable our method to generalize to navigator volumes interleaved with diagnostic slices, we additionally simulate a spin history artifact from the preceding slice~\cite{miller2015, gagoski2016}, which appears as dark shading along the slice imaging plane (blue arrows in Fig.~\ref{fig:examples}). We model this artifact with a Gaussian approximation of the slice profile $\tilde{f}(x)\!=\!f(x)(1-\mathcal{N}((x-c_{\text{slice}})^Tn_{\text{slice}}; 0, \sigma^2)$, where $f$ and $\tilde{f}$ are the original and augmented volumes, respectively; $c_{\text{slice}}$ and $n_{\text{slice}}$ are the center and normal to the slice imaging plane, sampled uniformly at random; and $\sigma^2$ is determined by slice thickness and the rate of spin decay~\cite{kuklisovamurgasova2012} (see Appendix~\ref{sec:training} for more details).

\section{Experiments and Results}

\subsection{Data}

We train two variants of E(3)-Pose on two high-quality \textit{research} data\-sets. We first evaluate on in-distribution held-out test data. Then, we evaluate on two \textit{clinical} datasets to assess domain generalization and potential for clinical translation. All volumes are annotated for GT pose and brain/eyes segmentations (Appendix~\ref{sec:datasets}). For all in-house datasets, all volunteers consented to data collection, and the approval of all ethical and experimental procedures and protocols was granted by the Institutional Review Board of the Boston Children's Hospital (BCH).

\subsubsection{Research.} We train one instance of E(3)-Pose on \textbf{Research-Fetal}, a dataset of 3D whole-uterus MRI volumes in 153 pregnant volunteers acquired at BCH. We train another instance of our method on \textbf{dHCP}, a publicly available dataset of fetal brain MRI scans in 245 volunteers~\cite{karolis2025}. Fetus gestational age (GA) is 18-38 and 20-38 weeks in Research-Fetal and dHCP, respectively. All volumes have 2-3mm isotropic voxels. Since they are not interleaved with diagnostic slices, they do not contain spin history artifacts. We use training/validation/testing splits of 114/15/25 and 148/48/49 participants for Res\-earch-Fetal and dHCP, respectively. dHCP consists of higher quality volumes, presenting a more challenging clinical domain generalization task compared to Research-Fetal. Additionally, dHCP volumes are neatly cropped around the fetal head. Therefore, segmentation networks trained on dHCP fail to accurately predict the brain mask in volumes where the field of view is larger and includes uterine regions outside the head and maternal tissues (Fig.~\ref{fig:examples}). For this reason, we train our segmentation network only on Research-Fetal.

\subsubsection{Clinical.} We evaluate the methods on \textbf{Clinical-Young}, a dataset of clinical volumes in 60 \textit{younger} fetuses (GA 18-23 weeks). This dataset presents two challenges: clinical volumes exhibit a substantial domain shift from high-quality research data, and underdeveloped anatomy complicates pose estimation~\cite{zhan2013}. The voxel size is 1.76-3.5mm, and volumes are not interleaved with diagnostic slices. We also test the models on \textbf{Navigators}, a dataset containing 47 time-series (1210 total volumes) of 3D MRI navigator volumes interleaved with 2D slices in 9 volunteers (GA 26-36 weeks). Due to fast and interleaved acquisitions, these volumes have voxel size 4-6mm and contain real spin history artifacts. This dataset represents the largest domain gap from the research data, and volumes are representative of the intended clinical application, i.e, adaptive slice prescription. Both clinical datasets are acquired at BCH.

\subsection{Evaluation Metrics}

We assess performance with the geodesic rotation error $\arccos(\frac{1}{2}[\text{trace}(R\hat{R}^{-1}) $\allowbreak$ -1])$ and the average absolute distance (AAD) for voxels on the brain surface~\cite{greve2009} (see Appendix~\ref{sec:additional_results} for additional metrics).

\subsection{Baseline Methods}

\subsubsection{Template-based.} We evaluate two baseline methods that perform rigid registration to a template volume. FireANTs~\cite{jena2024} is a state-of-the-art method for fast, optimization-based 3D image registration. EquiTrack~\cite{billot2024} uses a SE(3)-equivariant network that learns matching landmarks to compute the optimal rigid transform from the input to the template via SVD~\cite{levinson2020}. To enable comparison with E(3)-Pose, we generate subject-specific template volumes using the GT pose. For accurate registration, we mask input volumes with the brain masks predicted by our segmentation network. 

\subsubsection{Template-free.} We evaluate four baseline methods that estimate the canonical pose from the input volume without registering it to a template. Fetal-Align~\cite{hoffmann2021} implements landmark-based pose estimation, which segments the brain and eyes as landmarks and uses brain shape to resolve left-right ambiguity. For fair comparison on our datasets, we train our segmentation network to segment both the brain and eyes and use it to support landmark detection. The remaining three baselines directly regress pose with CNNs. 3DPose-Net~\cite{salehi2019} and 6DRep~\cite{faghihpirayesh2023} regress the axis-angle vector and two basis directions of the object frame, respectively. The latter method applies Gram-Schmidt orthonormalization~\cite{bjorck1967} to the network output to obtain the final rotation. Registration-by-Regression (RbR)~\cite{gopinath2024} trains a U-Net to regress a deformation field of the object frame at the resolution of the input volume. The final rotation is computed from the network output with SVD~\cite{horn_svd_1987}. We train all three networks on the research datasets and tune augmentation parameters for optimal performance. We augment these rotation estimation methods with CoM-based translation estimation to provide full rigid pose estimation. Appendix~\ref{sec:baselines} provides further details on the baseline methods.

\subsection{Results}

\begin{figure*}[t]
\centering
\centerline{\includegraphics[width=\textwidth]{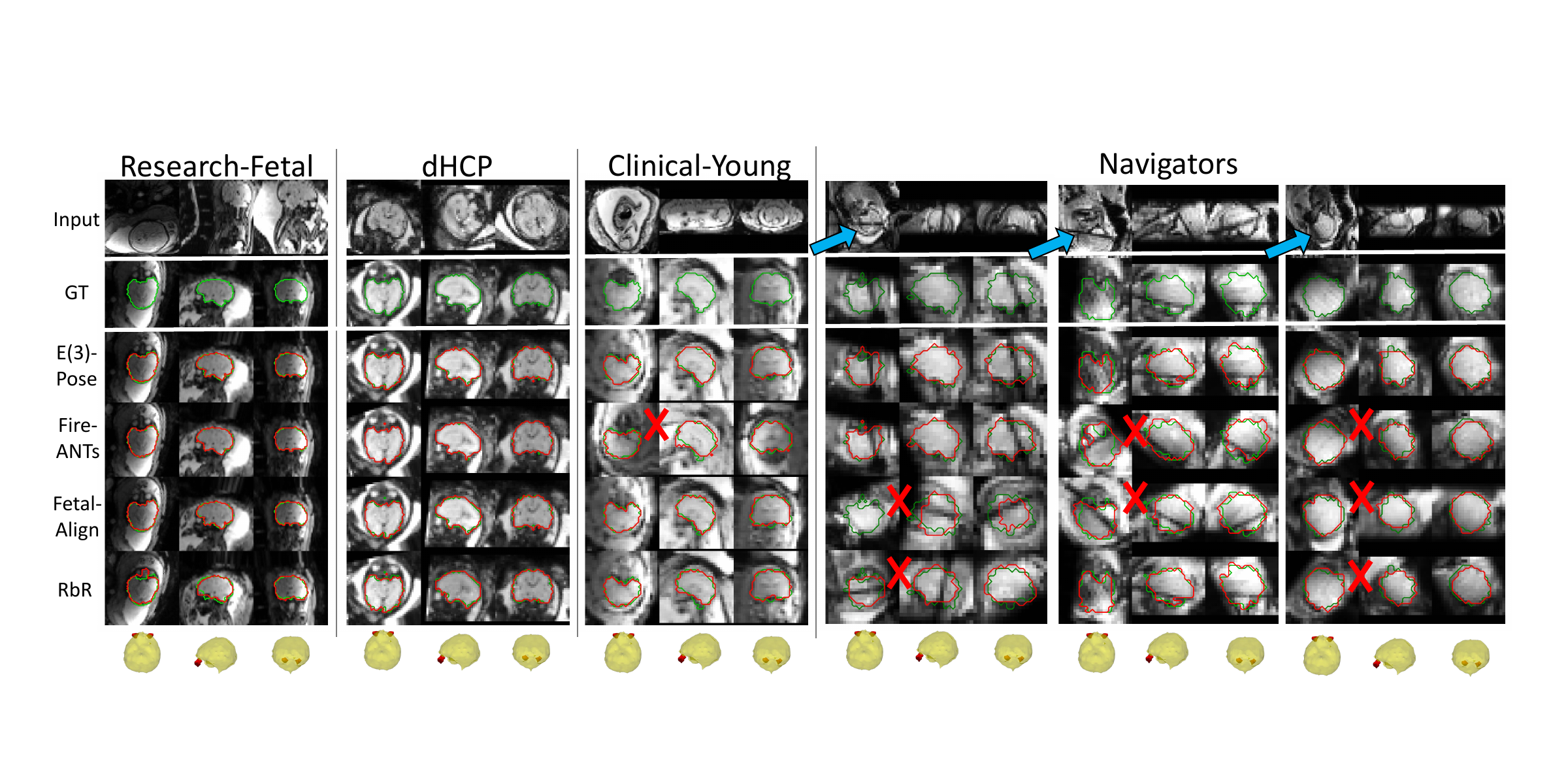}}
\caption{\textbf{Example results for methods trained on Research-Fetal.} Volumes are displayed before (row 1) and after (rows 2-6) alignment to the canonical object frame. The GT brain mask is also aligned to the GT (green outline) and predicted (red outline) frames. Navigator volumes include spin history artifacts (blue arrows) and low resolution/SNR, posing challenges for pose estimation. While baseline methods struggle (red Xs, rotation error $\!>60^\circ\!$) in younger fetuses (column 3) and navigator volumes (columns 4-6), E(3)-Pose correctly predicts pose in all cases. E(3)-Pose remains accurate under significant pose ambiguity, e.g. when the artifact intersects both eyes, large voxel size obscures brain structure, and the fetal brain is close to a sphere (column 6). See Figs.\ref{fig:examples_methods}-\ref{fig:examples_ablations} in Appendix~\ref{sec:additional_results} for additional examples.}
\label{fig:examples}
\end{figure*}

We first evaluate performance on in-domain research data, and then assess generalization to target clinical domains. Table~\ref{tab:baselines} reports the performance statistics on the test volumes from each dataset, and Fig.~\ref{fig:examples} shows example alignments. 

\begin{table*}[t]
\centering
\setlength\tabcolsep{0pt}
\fontsize{7}{8.5}\selectfont
\caption{\textbf{Baseline Comparisons.}  Mean $\pm$ standard deviation for rotation error ($^\circ$) and average absolute error (AAD, mm) are reported. We separately report the performance statistics for FireANTs and Fetal-Align, which do not rely on training (beyond segmentation). Best score under each training setting is shown in bold. Best overall score is italicized. * indicates statistical significance compared to E(3)-Pose ($p\!<\!0.05$, Bonferroni correction). See Appendix~\ref{sec:additional_results} for details on statistical testing. FireANTs runtime is 1.8s, prohibiting real-time deployment. EquiTrack runtime is 0.7s, and the remaining methods run in 0.3s total (including the segmentation step). E(3)-Pose outperforms all baselines on out-of-distribution, clinically representative datasets across training environments.}

\newcommand{\pmsp}{\,$\pm$\,}%
\resizebox{\textwidth}{!}{%
  \begin{tabular}{@{}l@{\;}
  r@{\,\(\pm\)\,}l@{\;}r@{\,\(\pm\)\,}l
  r@{\,\(\pm\)\,}l@{\;}r@{\,\(\pm\)\,}l
  r@{\,\(\pm\)\,}l@{\;}r@{\,\(\pm\)\,}l
  r@{\,\(\pm\)\,}l@{\;}r@{\,\(\pm\)\,}l@{}}
\toprule
& \multicolumn{4}{c}{\textbf{Research-Fetal test}}
& \multicolumn{4}{c}{\textbf{dHCP test}}
& \multicolumn{4}{c}{\textbf{Clinical-Young}}
& \multicolumn{4}{c}{\textbf{Navigators}} \\
\cmidrule(lr){2-5}\cmidrule(lr){6-9}\cmidrule(lr){10-13}\cmidrule(lr){14-17}
& \multicolumn{2}{c}{Rot. err.} & \multicolumn{2}{c}{AAD}
& \multicolumn{2}{c}{Rot. err.} & \multicolumn{2}{c}{AAD}
& \multicolumn{2}{c}{Rot. err.} & \multicolumn{2}{c}{AAD}
& \multicolumn{2}{c}{Rot. err.} & \multicolumn{2}{c}{AAD} \\
\midrule
\multicolumn{17}{l}{\textit{No training}} \\
\midrule
FireANTs~\cite{jena2024}
  & $10.6$ & $24.5$  & $5.7$ & $9.3^*$
  & $\textit{\textbf{0.4}}$ & $\textit{\textbf{1.6}}^*$  & $\textit{\textbf{0.6}}$ & $\textit{\textbf{1.0}}^*$
  & $17.3$ & $40.6$    & $4.1$ & $7.4$
  & $\mathbf{44.6}$ & $\mathbf{55.6^*}$  & $\mathbf{19.1}$ & $\mathbf{20.3^*}$ \\
Fetal-Align~\cite{hoffmann2021}
  & $\textit{\textbf{5.0}}$ & $\textit{\textbf{3.0}}$  & $\textit{\textbf{3.0}}$ & $\textit{\textbf{1.7}}$
  & $8.5$ & $24.7^*$  & $3.5$ & $6.8^*$
  & $\mathbf{10.3}$ & $\mathbf{21.0^*}$  & $\mathbf{3.2}$ & $\mathbf{5.0^*}$
  & $56.2$ & $57.6^*$  & $24.1$ & $23.3^*$ \\
\midrule
\multicolumn{17}{l}{\textit{Trained on Research-Fetal}} \\
\midrule
EquiTrack~\cite{billot2024}
  & $11.3$ & $13.1^*$  & $6.2$ & $5.9^*$
  & $\mathbf{7.1}$ & $\mathbf{5.0}$     & $\mathbf{3.6}$ & $\mathbf{2.6}$
  & $18.9$ & $29.0^*$  & $5.4$ & $5.7^*$
  & $44.7$ & $48.7^*$  & $20.3$ & $18.3^*$ \\
3DPose-Net~\cite{salehi2019}
  & $19.8$ & $15.0^*$  & $10.5$ & $7.6^*$
  & $24.0$ & $19.2^*$  & $10.9$ & $7.5^*$
  & $39.2$ & $36.8^*$  & $11.2$ & $8.6^*$
  & $65.0$ & $47.8^*$  & $27.7$ & $16.3^*$ \\
6DRep~\cite{faghihpirayesh2023}
  & $9.6$ & $4.5^*$   & $5.3$ & $2.3^*$
  & $9.8$ & $5.5^*$   & $4.9$ & $3.0^*$
  & $14.1$ & $7.7^*$  & $4.6$ & $2.3^*$
  & $38.8$ & $45.7^*$ & $17.2$ & $15.9^*$ \\
RbR~\cite{gopinath2024}
  & $7.6$ & $3.9^*$   & $4.3$ & $2.2^*$
  & $8.6$ & $5.0$     & $4.3$ & $2.6$
  & $11.2$ & $4.6^*$  & $3.7$ & $1.4^*$
  & $22.6$ & $24.1^*$ & $11.5$ & $9.9^*$ \\
E(3)-Pose (ours)
  & $\mathbf{5.1}$ & $\mathbf{2.6}$  & $\mathbf{3.0}$ & $\mathbf{1.7}$
  & $7.4$ & $3.6$     & $3.7$ & $1.8$
  & $\textit{\textbf{9.1}}$ & $\textit{\textbf{4.7}}$  & $\textit{\textbf{3.0}}$ & $\textit{\textbf{1.4}}$
  & $\textit{\textbf{9.4}}$ & $\textit{\textbf{7.5}}$  & $\textit{\textbf{6.3}}$ & $\textit{\textbf{4.0}}$ \\
\midrule
\multicolumn{17}{l}{\textit{Trained on dHCP}} \\
\midrule
EquiTrack~\cite{billot2024}
  & $29.5$ & $43.5^*$  & $14.1$ & $19.5^*$
  & $13.6$ & $36.7^*$  & $5.4$ & $11.6^*$
  & $48.3$ & $62.1^*$  & $11.6$ & $13.2^*$
  & $59.0$ & $57.5^*$  & $25.7$ & $22.1^*$ \\
3DPose-Net~\cite{salehi2019}
  & $33.5$ & $26.6^*$  & $17.1$ & $11.6^*$
  & $21.2$ & $27.4^*$  & $9.0$ & $8.9^*$
  & $51.5$ & $37.8^*$  & $14.4$ & $8.9^*$
  & $93.9$ & $55.8^*$  & $36.2$ & $16.7^*$ \\
6DRep~\cite{faghihpirayesh2023}
  & $32.2$ & $26.0^*$  & $17.3$ & $14.2^*$
  & $12.2$ & $6.1^*$   & $5.9$ & $3.1^*$
  & $48.2$ & $50.8^*$  & $12.4$ & $10.8^*$
  & $84.1$ & $51.2^*$  & $34.0$ & $17.3^*$ \\
RbR~\cite{gopinath2024}
  & $15.4$ & $18.3^*$  & $8.3$ & $8.5^*$
  & $8.0$ & $4.0^*$   & $3.9$ & $1.9$
  & $50.9$ & $48.4^*$  & $12.7$ & $9.9^*$
  & $77.2$ & $53.6^*$  & $30.8$ & $17.3^*$ \\
E(3)-Pose (ours)
  & $\mathbf{5.7}$ & $\mathbf{3.6}$  & $\mathbf{3.4}$ & $\mathbf{2.0}$
  & $\mathbf{7.3}$ & $\mathbf{3.4}$  & $\mathbf{3.7}$ & $\mathbf{1.7}$
  & $\mathbf{12.1}$ & $\mathbf{6.8}$ & $\mathbf{3.9}$ & $\mathbf{1.9}$
  & $\mathbf{13.9}$ & $\mathbf{13.2}$& $\mathbf{7.9}$ & $\mathbf{5.6}$ \\
\bottomrule
\end{tabular}%
}
\label{tab:baselines}
\end{table*}

\subsubsection{Research.}\label{sec:results_research}
E(3)-Pose accurately predicts pose and significantly outperforms other direct pose regression methods, showcasing the advantage of modeling rotational equivariance. The relative rankings of 3DPose-Net, 6DRep, and RbR highlight the stability gains of estimating continuous and higher-dimensional rotation parametrizations~\cite{zhou2019}. E(3)-Pose takes advantage of this concept by regressing a continuous, overcomplete output. On Research-Fetal test volumes, Fetal-Align demonstrates similar performance to E(3)-Pose, indicating that precise eye segmentation supports reliable pose estimation that is consistent with our landmark-based GT pose definition (Appendix~\ref{sec:annotations}). E(3)-Pose is outperformed by template-based methods on dHCP test volumes, where the high-quality intensity distributions of the input and template are almost the same, eliminating alignment ambiguities. As a result, these methods effectively exploit the (unfair) information advantage of a subject-specific template already aligned to the GT pose. Furthermore, the 1.8s inference time of FireANTs (c.f. 0.3s for E(3)-Pose) prohibits clinical deployment.

\subsubsection{Clinical-Young.} E(3)-Pose yields accurate pose estimates in clinical acquisitions of younger ages, outperforming all baselines. This difference is more pronounced when training on dHCP, which consists of higher-SNR volumes and presents a larger distribution shift from clinical data compared to Research-Fetal. E(3)-Pose is the only method that maintains its performance in both training settings, highlighting the robust generalization advantages of our method. For further insight, we study the performance as a function of GA. Fig.~\ref{fig:early_ga}a reveals that template-based methods (FireANTs and EquiTrack) break down on younger fetuses, where underdeveloped, spherical brain shapes and the lack of structure differentiation within the brain lead to alignment ambiguities~\cite{zhan2013}. When trained on Research-Fetal, direct pose regression (6DRep and RbR) remains relatively stable across GAs. By removing dependence on specific anatomical markers, these methods are less sensitive to the lack of fine-grain details. However, when trained on dHCP, standard CNNs struggle to generalize to younger GAs, suggesting that these methods are prone to overfitting. In contrast, E(3)-Pose provides accurate results across GAs and training datasets.

\subsubsection{Navigators.} E(3)-Pose achieves accurate pose estimation in navigator volumes and significantly outperforms all baselines. Template registration methods are prone to errors when spin history artifacts disrupt the similarity between input and template intensity distributions, and Fetal-Align remains highly sensitive to eye visibility, which is often limited by artifact obstruction or blurring due to voxel size. Furthermore, all baselines that rely on training struggle with the challenging task of generalizing from dHCP to Navigators. In contrast, E(3)-Pose remains robust to the substantial domain gap. Importantly, E(3)-Pose successfully recovers head pose even in volumes with high uncertainty, which occur when the effects of low resolution, low SNR, and artifact obstruction combine to eliminate most asymmetric cues (Fig.~\ref{fig:examples}, columns 4-6). Moreover, E(3)-Pose generally yields more stable estimates compared to CNN-based pose regression, highlighting that enforcing pose equivariance and anatomical symmetry on the network parameters is an effective regularization technique that enables generalization to out-of-distribution clinical volumes and stability under pose ambiguity.

\subsection{Ablation Study}

To study performance on challenging out-of-distribution cases, we evaluate ablations on clinical datasets (Table~\ref{tab:ablation}). Appendix~\ref{sec:ablation_supp} provides further evaluations.

\subsubsection{Equivariance.} We ablate the modeling of rotational equivariance by replacing all E(3)-equivariant convolutions with standard convolutions equivariant under translations only. The results show that rotational equivariance is critical for generalization to clinical data, especially when the domain gap between training and testing data is significant. This outcome is in line with prior work~\cite{lawrence2022, sannai2019, bulusu2022} and our comparison to non-equivariant RbR, 6DRep, and 3DPose-Net.

\begin{table*}[t]
\centering
\setlength\tabcolsep{0pt}
\fontsize{7}{8.5}\selectfont
\caption{\textbf{Ablation Study.} Mean $\pm$ standard deviation statistics for rotation error ($^\circ$) and average absolute error (AAD, mm) are reported. Best score is shown in bold. * indicates statistical significance compared to E(3)-Pose ($p\!<\!0.05$, Bonferroni correction). See Appendix~\ref{sec:additional_results} for details on statistical testing and subject-level statistics in Navigators.}

\resizebox{\textwidth}{!}{%
\begin{tabular}{@{}l@{\;}
  r@{\,\(\pm\)\,}l@{\;}r@{\,\(\pm\)\,}l
  r@{\,\(\pm\)\,}l@{\;}r@{\,\(\pm\)\,}l
  r@{\,\(\pm\)\,}l@{\;}r@{\,\(\pm\)\,}l
  r@{\,\(\pm\)\,}l@{\;}r@{\,\(\pm\)\,}l@{}}
\toprule
& \multicolumn{8}{c}{\textbf{Trained on Research-Fetal}}
& \multicolumn{8}{c}{\textbf{Trained on dHCP}} \\
\cmidrule(lr){2-9}\cmidrule(lr){10-17}
& \multicolumn{4}{c}{\textbf{Clinical-Young}} & \multicolumn{4}{c}{\textbf{Navigators}}
& \multicolumn{4}{c}{\textbf{Clinical-Young}} & \multicolumn{4}{c}{\textbf{Navigators}} \\
\cmidrule(lr){2-5}\cmidrule(lr){6-9}\cmidrule(lr){10-13}\cmidrule(lr){14-17}
& \multicolumn{2}{c}{Rot. err.} & \multicolumn{2}{c}{AAD}
& \multicolumn{2}{c}{Rot. err.} & \multicolumn{2}{c}{AAD}
& \multicolumn{2}{c}{Rot. err.} & \multicolumn{2}{c}{AAD}
& \multicolumn{2}{c}{Rot. err.} & \multicolumn{2}{c}{AAD} \\
\midrule
E(3)-Pose (ours)
  & $9.1$ & $4.7$
  & $\mathbf{3.0}$ & $\mathbf{1.4}$
  & $\mathbf{9.4}$ & $\mathbf{7.5}$
  & $\mathbf{6.3}$ & $\mathbf{4.0}$
  & $12.1$ & $6.8$
  & $3.9$ & $1.9$
  & $\mathbf{13.9}$ & $\mathbf{13.2}$
  & $\mathbf{7.9}$ & $\mathbf{5.6}$ \\
\midrule
Standard CNN
  & $10.3$ & $5.4^*$
  & $3.4$ & $1.5^*$
  & $18.0$ & $19.3^*$
  & $9.7$ & $7.6^*$
  & $53.3$ & $55.8^*$
  & $12.9$ & $11.9^*$
  & $80.2$ & $52.7^*$
  & $31.9$ & $17.3^*$ \\
\midrule
No pseudovector
  & $10.4$ & $4.0^*$
  & $3.4$ & $1.3^*$
  & $10.9$ & $11.1$
  & $7.0$ & $5.0$
  & $12.1$ & $7.7$
  & $3.9$ & $2.2$
  & $20.6$ & $26.1^*$
  & $10.6$ & $10.1^*$ \\
$h(R)\!=\!e_y\!\oplus\!e_z$
  & $10.3$ & $4.2^*$
  & $3.4$ & $1.3^*$
  & $11.9$ & $12.9$
  & $7.3$ & $5.5^*$
  & $11.6$ & $6.5$
  & $3.7$ & $1.8$
  & $26.1$ & $35.6^*$
  & $12.3$ & $12.2^*$ \\
\midrule
$|\sin{\frac{\theta_x}{2}}|$
  & $\mathbf{8.8}$ & $\mathbf{3.6}$
  & $\mathbf{3.0}$ & $\mathbf{1.3}$
  & $12.1$ & $11.1^*$
  & $7.2$ & $5.1^*$
  & $\mathbf{11.1}$ & $\mathbf{5.3}$
  & $\mathbf{3.6}$ & $\mathbf{1.5}$
  & $18.5$ & $27.3$
  & $9.4$ & $9.4$ \\
Geodesic loss
  & $9.2$ & $4.7$
  & $3.1$ & $1.5$
  & $13.5$ & $20.0$
  & $9.7$ & $7.6$
  & $30.9$ & $53.4$
  & $7.7$ & $10.9$
  & $27.8$ & $42.5^*$
  & $12.6$ & $14.0^*$ \\
No artifact augm.
  & $12.0$ & $21.1$
  & $3.5$ & $3.8$
  & $16.2$ & $18.6^*$
  & $7.7$ & $7.6^*$
  & $18.4$ & $21.1^*$
  & $5.4$ & $4.5^*$
  & $31.9$ & $36.7^*$
  & $15.1$ & $13.9^*$ \\
\bottomrule
\end{tabular}%
}
\label{tab:ablation}
\end{table*}

\begin{figure}[t]
\centering
\centerline{\includegraphics[width=0.8\textwidth]{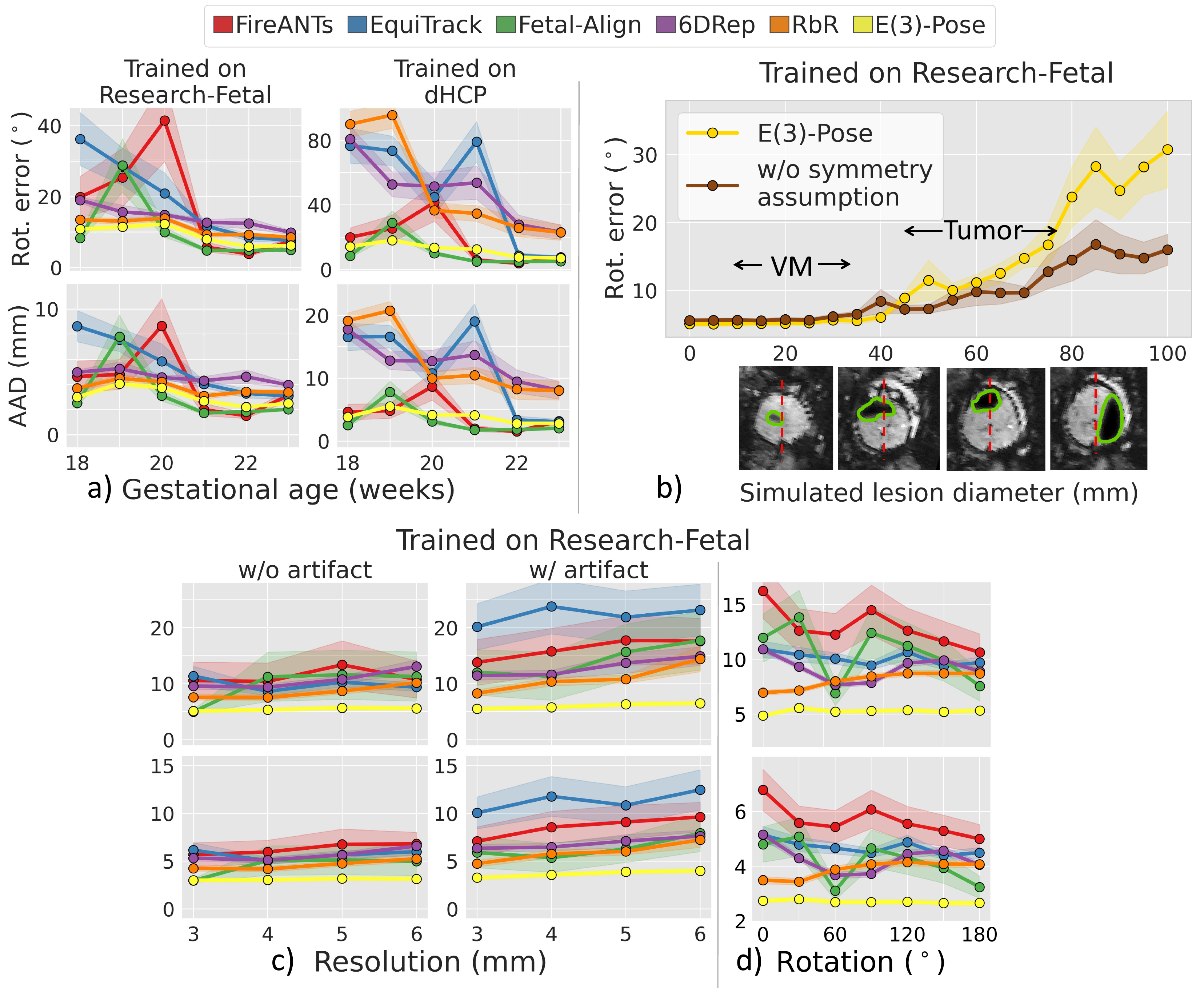}}
\caption{\textbf{Fine-grain analysis of performance.} a) Performance statistics on Clinical-Young as a function of gestational age for all well-performing methods. b) E(3)-Pose produces stable pose estimates in test volumes from Research-Fetal with simulated asymmetric lesions (green outline) comparable in size to ventriculomegaly (VM) of 10-30mm ~\cite{alluhaybi2022}, but breaks down under large asymmetries comparable to tumors of 40-70mm~\cite{bedei2022}, where the asymmetric variant of our method represents a suitable alternative strategy. c) E(3)-Pose remains robust to progressively larger voxel sizes (\textit{left}), even when simulating additional spin history artifacts (\textit{right}) in augmented test volumes from Research-Fetal. E(3)-Pose significantly outperforms all baseline methods on 6mm volumes with artifacts ($p\!<\!0.05$, Bonferroni-corrected pairwise Wilcoxon).  d) E(3)-Pose produces stable pose estimates across a wide range of input poses in rotated test volumes from Research-Fetal.
}
\label{fig:early_ga}
\end{figure}

\subsubsection{Rotation Parametrization.}We consider alternative equivariant rotation par-ametrizations. First, we study the impact of modeling left-right head symmetry with a pseudovector by replacing it with a standard vector. E(3)-Pose significantly outperforms this ablation, which struggles under pose ambiguities. Second, we reduce the dimensionality of the rotation parametrization by predicting only two basis directions: $e_y$ and $e_z$. We observe that estimating a third basis direction significantly reduces errors in challenging clinical volumes.

\subsubsection{Loss Function and Data Augmentation.} We examine the effect of two alternative loss functions (Appendix~\ref{sec:ablation_supp}). We first train our network by replacing the pseudovector loss term in Eq. (\ref{eq:loss}) with that of the standard vectors. Next, we replace our objective with the widely used geodesic loss~\cite{salehi2019}. Both loss functions significantly reduce performance in Navigators, emphasizing the importance of combining both symmetry-aware architectures and loss functions for stable performance under high ambiguity. Lastly, ablating the simulated spin history artifact during training shows that this augmentation critically enables generalization to Navigators, where volumes contain real spin history artifacts.

\subsection{Sensitivity Study }

\label{sec:sensitivity}

\subsubsection{Symmetry Perturbations}

To investigate the scope of our symmetry assumption under pathological asymmetries, we simulate lateralized lesions in test volumes from Research-Fetal (Fig.~\ref{fig:early_ga}b). E(3)-Pose is robust to simulated asymmetries that are comparable in size to most observed pathologies. Under severe asymmetries, ablating the symmetric construction of the rotation parametrization supports more accurate pose estimation. 

\subsubsection{Intensity Perturbations.}
To understand performance gaps between Research-Fetal and Navigators, we assess robustness to simulated larger voxel size and spin history artifacts in test volumes from Research-Fetal (Fig.~\ref{fig:early_ga}c). E(3)-Pose remains much more consistent than all five tested baseline algorithms, which become increasingly error-prone with increasingly severe perturbations. E(3)-Pose exhibits the smallest increase in both error metrics from the original 3mm volumes to the 6mm volumes with simulated spin history artifacts and outperforms all baselines on the latter, most notably compared to template-dependent methods, which are highly sensitive to the presence of disruptive artifacts. 

\subsubsection{Spatial Perturbations.}
To evaluate robustness of the method across input poses, we use test volumes from Research-Fetal to simulate a wide range of rotational misalignments of the canonical object frame relative to the input volume. E(3)-Pose displays the most stable consistency across rotations (Fig.~\ref{fig:early_ga}d), underscoring the value of modeling rotation equivariance in pose estimation.

\begin{figure*}
\centering
\centerline{\includegraphics[width=\textwidth]{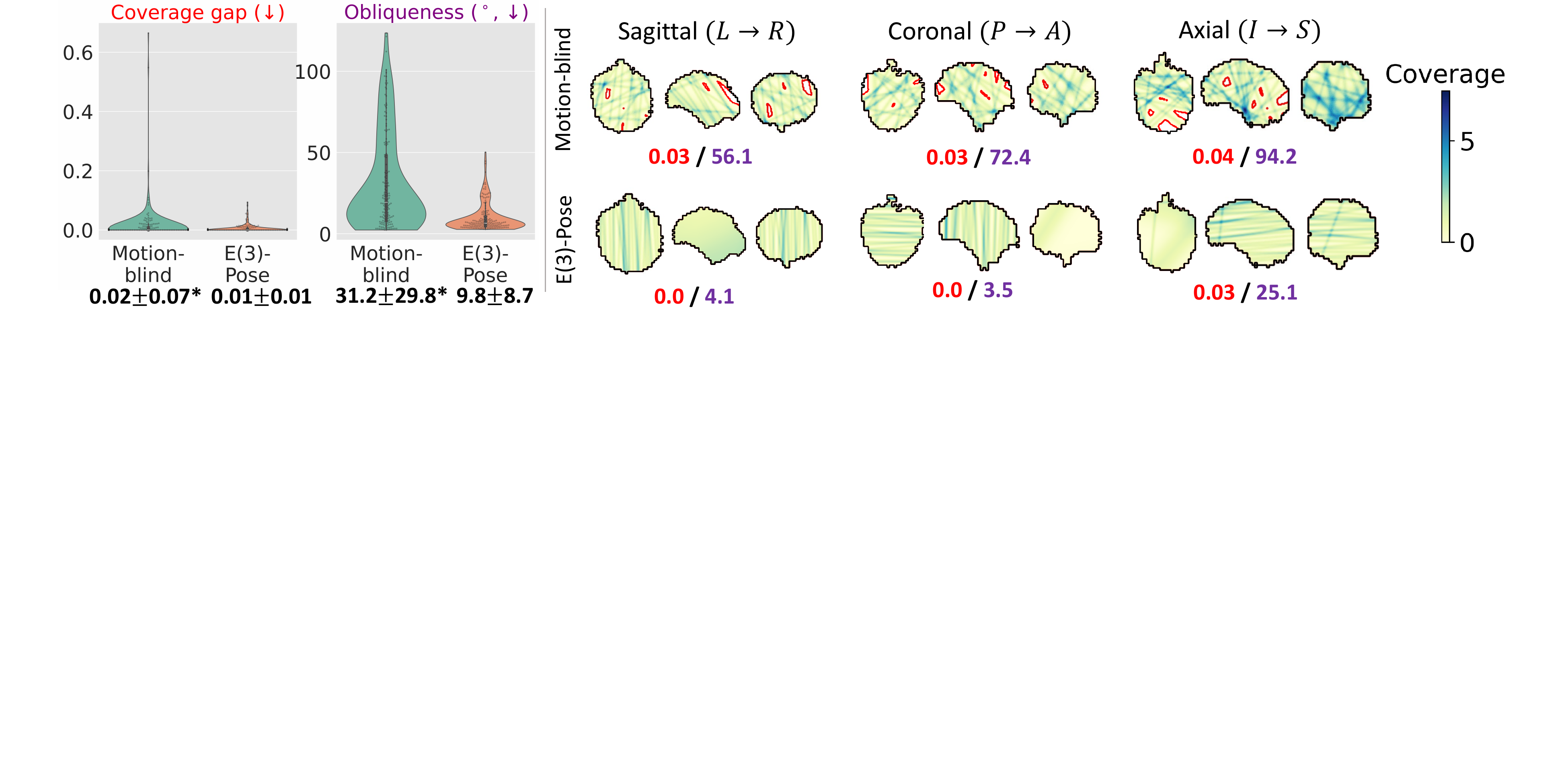}}
\caption{\textbf{Simulation results.} \textit{Left:} Quantitative comparison of diagnostic slice stacks obtained using motion-blind prescription and E(3)-Pose. Mean $\pm$ standard deviation statistics are reported. * indicates statistical significance ($p\!<\!0.05$, pairwise Wilcoxon). \textit{Right:} Brain coverage of the slice stacks prescribed by each method, for three different example subjects and target anatomical orientations. Coverage gap (red) and obliqueness (purple, $^\circ$) are respectively displayed. Spatial coverage gaps are outlined in red.}
\label{fig:sim}
\end{figure*}

\subsection{Simulation Study}
We further evaluate the potential of E(3)-Pose to support automated slice prescription through simulations using test volumes from Research-Fetal. We simulate stacks of slices in the sagittal (L$\rightarrow$R), coronal (P$\rightarrow$A), and axial (I$\rightarrow$S) orientations in all subjects under real fetal motion trajectories. We compare our method to a ``motion-blind'' strategy, which prescribes every slice based on the head pose in the first navigator volume only. We assess the quality of the simulated stacks by computing the coverage gap (i.e., the ratio of brain volume that is not captured) and slice obliqueness (i.e., rotation error between the prescribed and target slice orientations). Appendix~\ref{sec:sim} provides implementation details and additional evaluations. Fig.~\ref{fig:sim} shows that E(3)-Pose yields significant improvements in both metrics. Under motion-blind prescription, large inter-slice motion leads to oblique slices that frequently intersect, producing highly uneven spatial coverage and, in the worst cases, complete coverage gaps. In contrast, E(3)-Pose adaptively adjusts the imaging plane to follow the movements of the head. Our approach prescribes parallel slices closely aligned to the target anatomical orientation, improving the uniformity of coverage and reducing the likelihood of coverage gaps in our simulations.

\section{Discussion and Conclusion}

We present E(3)-Pose, a novel framework for 6-DoF pose estimation. By jointly modeling rigid pose equivariance and reflectional object symmetry, E(3)-Pose enables robust generalization to challenging domains with pose ambiguities.  

Under large lateralized pathologies, our symmetry assumption, which is only an approximation based on the low resolution of navigators, may not hold. We validate the asymmetric variant of E(3)-Pose as a robust fallback strategy in simulated asymmetries, and we leave evaluation on real pathological cases to future work. In addition, E(3)-Pose does not explicitly model motion history or pose uncertainty, both of which could help to mitigate errors in noisy clinical volumes. Future work will investigate equivariant and symmetry-aware approaches for probabilistic reasoning over SO(3), conditioned on motion trajectories~\cite{klee2023,peretroukhin2020,rangaprasad2016}. See Appendix~\ref{sec:future_work} for further discussion.

In our experiments, we provide the first evaluation on data representative of clinical applications in fetal brain MRI, supporting the potential for future clinical deployment.

\section*{Acknowledgments}
This work is supported by the the MIT Abdul Latif Jameel Clinic, the MIT-Takeda Program, the Chou Family Transformative Research Fund, NIH NIBIB (R01EB032708, 5T32EB1680), and NIH NICHD (R01HD100009). This work involved human subjects or animals in its research. Approval of all ethical and experimental procedures and protocols was granted by the Institutional Review Board of Boston Children’s Hospital.

%
%

\clearpage
\appendix
\setcounter{page}{1}

\title{Supplementary Material}

\section{Automated Slice Prescription}
\label{sec:interleaved}

In this section, we describe our clinical application in detail.

\subsubsection{Overview.} Radiological assessment of abnormal fetal brain development relies on volumetric fetal brain MRI, which provides greater contrast and resolution than standard ultrasound examinations. Due to the high sensitivity of 3D MRI acquisitons to fetal motion, current clinical practice acquires stacks of diagnostic-quality 2D MRI slices in the three target anatomical directions (i.e., sagittal, coronal, and axial)~\cite{glenn2009}. Slices that closely align to the target anatomical orientation are critical for abnormality detection. For instance, non-oblique midline sagittal slices are necessary for detecting agenesis of the corpus callosum~\cite{gholipour2014}. Furthermore, inter-slice fetal motion is often correlated with poor spatial coverage. Even small coverage gaps can substantially reduce diagnostic potential: for example, in second- and third-trimester fetuses, the proportion of the total brain volume constituted by the cerebellum ranges from only 0.03 to 0.06~\cite{cai2020}. Adaptive slice prescription based on the fetal head motion thus holds potential to improve the quality of prenatal care.

\begin{figure}
\centering
\centerline{\includegraphics[width=0.8\textwidth]{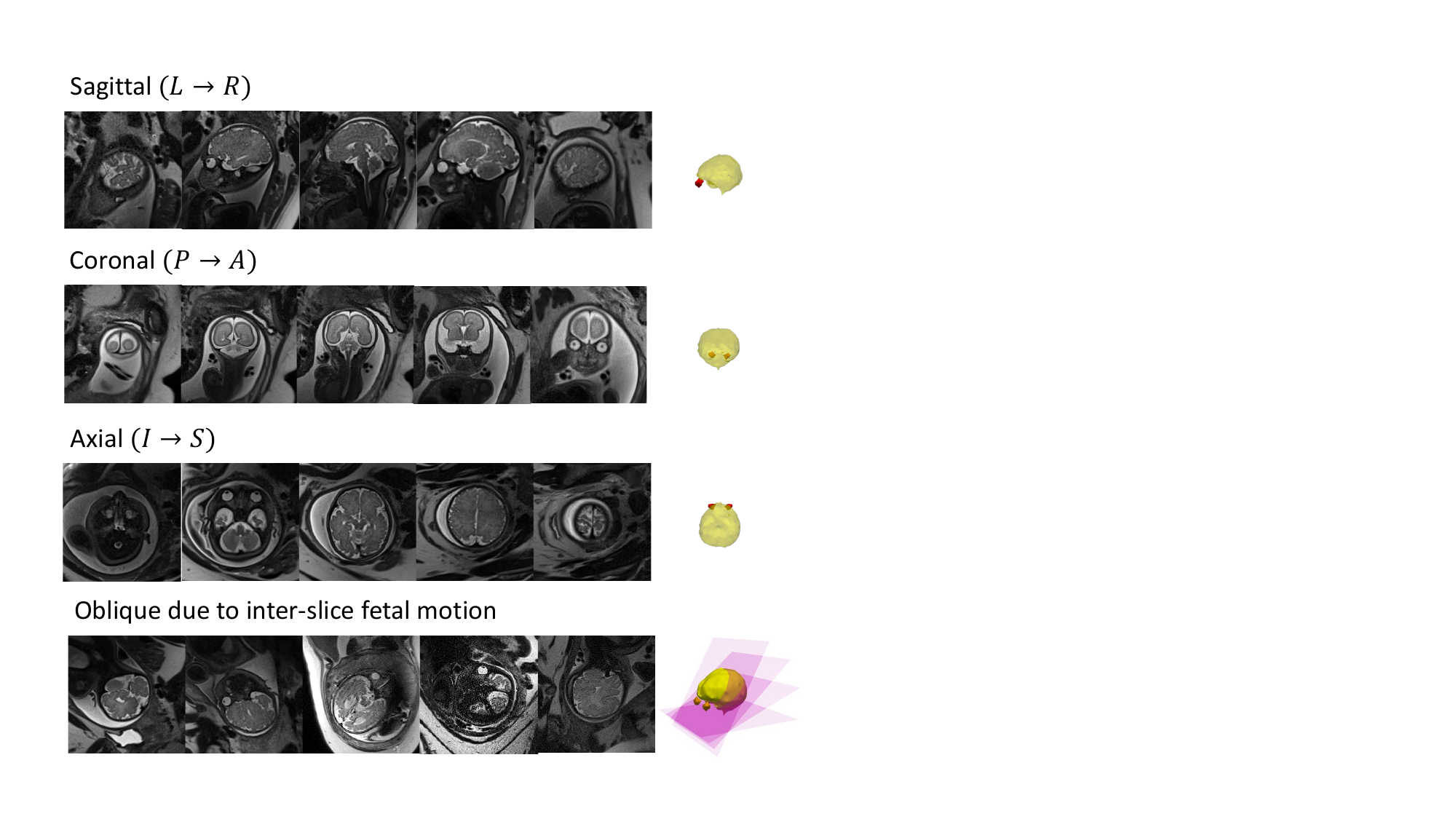}}
\caption{\textbf{Example diagnostic 2D slices.} Accurate radiological assessment of developmental abnormalities necessitates slices aligned to the sagittal (left$\rightarrow$right), coronal (posterior$\rightarrow$anterior), and axial (inferior$\rightarrow$superior) anatomical directions (\textit{rows 1-3}). Inter-slice fetal motion causes oblique slice orientations relative to the canonical anatomical directions (\textit{row 4}). This produces slices that are challenging for the radiologist to interpret and induces coverage gaps in the acquired stack of slices.
}
\label{fig:slices}
\end{figure}

Slices are T2-weighted and acquired with the half-Fourier acquisition single-shot turbo spin-echo (HASTE) MRI sequence ($1.25\!\times\!1.25$mm pixels, 3mm slice thickness, $320\!\times\!320$mm FOV, TR=2.5s, TE=106ms, GRAPPA R=2, Partial Fourier=5/8, $90^\circ$ flip angle). Fig.~\ref{fig:slices} shows examples of slices in the target anatomical orientations, as well as \textit{oblique} slices caused by fetal motion. To automatically prescribe the imaging plane of each slice according to the current fetal head pose, we rapidly acquire a 3D navigator volume before each slice~\cite{gagoski2016}. Navigator volumes are acquired with the echo-planar imaging (EPI) sequence (4-6mm isotropic voxels, $324\!\times\!324\!\times\!120$mm FOV, TR=29ms, TE=14ms, 5$^\circ$ flip angle). Low-resolution, low-energy navigator volumes are not of diagnostic quality but suffice for head pose estimation.

\begin{figure*}
\centering
\centerline{\includegraphics[width=0.97\textwidth]{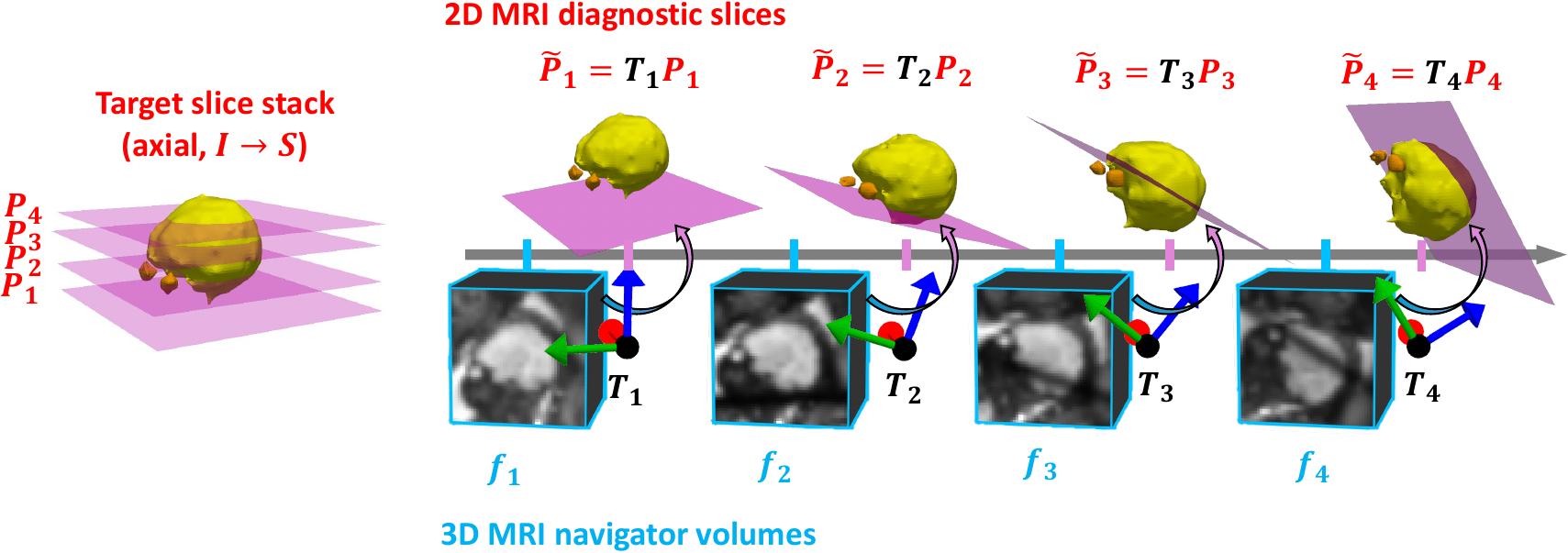}}
\caption{\textbf{Interleaved acquisition framework.} To correct for fetal head motion, we aim to automatically adjust the imaging plane $P_k$ of every 2D slice, using the current head pose $T_k$. We rapidly acquire a navigator volume $f_k$ before every slice in order to accurately estimate $T_k$ from $f_k$ with E(3)-Pose. 
}
\label{fig:interleaved}
\end{figure*}

\subsubsection{Automation Framework.} We aim to rapidly estimate the 6-DoF head pose in each navigator volume and use it to adjust the imaging plane of the next slice (Fig.~\ref{fig:interleaved}).  Formally, at the start of the sequence, we define the target imaging planes $P_1,...,P_K\!\in\!\text{SE(3)}$ of all $K$ slices in the stack, relative to the canonical fetal head frame. The rotational and translational components of $P_k$ indicate the target anatomical orientation and the relative position of the slice within the stack, respectively. Before the $k$th slice in the stack, we acquire a navigator volume, from which we rapidly estimate the head pose $T_k=t_k\circ R_k$ and prescribe the imaging plane of the next slice to be $\tilde{P}_k=T_kP_k$. We also translate the field-of-view (FOV) of the next navigator volume around the fetal head based on $t_k$. We include this step to ensure full head capture for accurate pose estimation. Because the time interval between every navigator and the next slice is 1s, we require the runtime of pose estimation to be less than 1s. 

\subsubsection{In-utero Implementation~\cite{muthukrishnan2026}.} 
\begin{figure*}
\centering
\centerline{\includegraphics[width=0.97\textwidth]{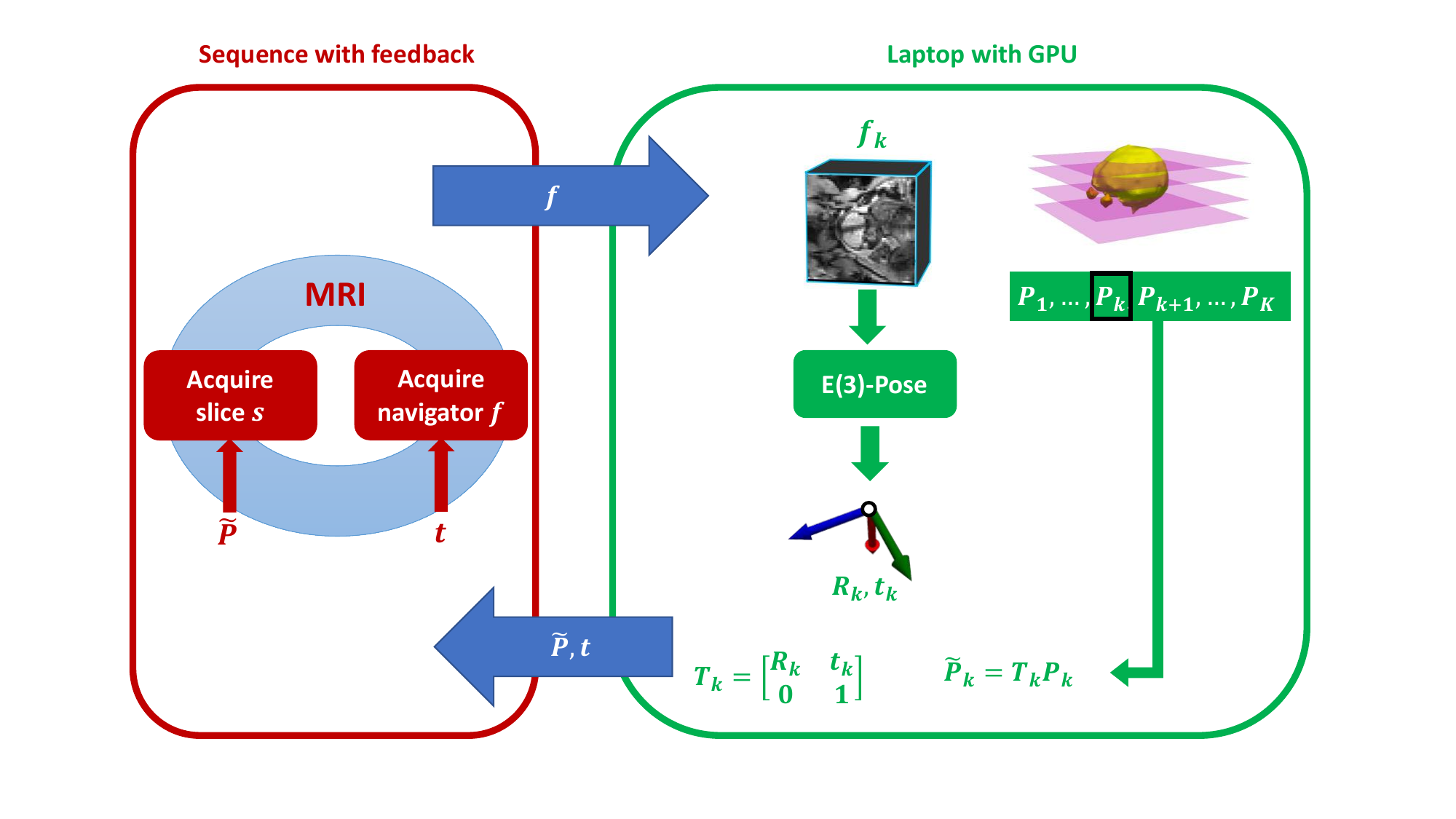}}
\caption{\textbf{Feedback loop system.} Our implementation runs on a 3T Siemens scanner, connected to a server hosted on a GPU-enabled laptop. On the server (\textit{right}), E(3)-Pose separately estimates the rotation $R_k$ and translation $t_k$ of the fetal head pose $T_k$, based on the navigator volume $f_k$ sent by the scanner (\textit{left}). We apply translation $t_k$ to center the next navigator volume FOV around the fetal head. Finally, we send the prescription parameters to the scanner for the next acquisition. Figure adapted from published work on our \textit{in utero} implementation~\cite{muthukrishnan2026}.
}
\label{fig:feedback}
\end{figure*}

We implemented our full automation framework on a GPU-enabled laptop server connected to a 3T Siemens scanner using existing software~\cite{wighton2024} (Fig.~\ref{fig:feedback}). While our published work on this implementation evaluates only automated FOV translation of the navigator volume FOV \textit{in utero}, we have since used the full pose returned by E(3)-Pose to automatically prescribe axial, sagittal, and coronal slice stacks \textit{in utero}. Future work will evaluate the diagnostic potential of prescribed slice stacks on a larger cohort of pregnant volunteers. In contrast to the published abstract and our ongoing work, which focus on clinical deployment only, the scope of this paper is the technical development of our method.

\section{E(3)-CNN Implementation Details}
\label{sec:e3cnn_arch}

\subsubsection{Equivariant Pooling and Non-linearities.} E(3)-CNNs necessitate special poolings and non-linearities, since standard versions only satisfy Eq.~(\ref{eq:equivariance}) for zero-order fields. Specifically, higher order ($l\!>\!0$) fields require (\textit{i})~norm-based pooling, and (\textit{ii})~special non-linearities learned by the network as multiplicative scalar fields: $f(x) \sigma(\kappa_{l,p,0,0}\!\ast\!f(x))$, where $\sigma$ is the sigmoid function, and $\ast$ is the convolution operator ~\cite{weiler2018}.

\subsubsection{Network Architecture.} We crop input volumes to the E(3)-CNN around the predicted brain mask to $64^3$ voxels, scaled such that the brain occupies $60\%$ of the cropped volume size. Our E(3)-CNN architecture has 4 levels, each consisting of 2 E(3)-equivariant convolutions~\cite{geiger2022,diaz2024} with kernel size $5\! \times\! 5\! \times \!5$, followed by instance normalization and equivariant max-pooling. Following~\cite{diaz2024}, the first level has 8, 4, and 2 even- and odd-parity output features of orders $l=0,1,2$, respectively, and the feature count is doubled in each level. We use ReLU and tanh activations for even and odd scalar features respectively, and equivariant sigmoid activations for higher-order features. For the radial basis functions $\varphi_m$ in the $k^\text{th}$ level, we use $\varphi_m(r)\!=\!8.433573\text{sus}(x+m-1)\text{sus}(1-m-x)$, where $\text{sus}$ is the soft unit step function implemented in~\cite{diaz2024} and parametrized by $m \in \{0, r/4, r/2, 3r/4, r\}$, where $r=2^{k-2}$. 

\section{Segmentation Network Architecture}
\label{sec:unet_arch}

Following~\cite{silva2024}, we implement the brain segmentation function $\psi$ with a standard 3D U-Net. Our architecture has 4 levels with 16, 32, 64, and 128 output channels, respectively, each consisting of 2 convolutional layers with kernel size $3\!\times\! 3\! \times\! 3$, followed by batch normalization and ELU activation~\cite{clevert2015}. Inputs are padded to $128^3$ voxels. While we only use the predicted brain mask for inference, we train the U-Net to additionally segment the eyes, with the aim of enhancing the overall robustness via multi-task learning~\cite{gao2020}. 

\section{E(3)-CNN Loss Function}
\label{sec:e3cnn_loss}

Let $e_k,\hat{e}_k\!\in\!\mathbb{R}^3$ be the GT and predicted basis directions, respectively, for $k\in \{x,y,z\}$, where $||e_k||_2\!=\!||\hat{e}_k||_2\!=\!1$ and $\theta_k\!=\!\arccos{(\hat{e}_k\cdot e_k)}$.

\begin{lemma}
$|\sin{(\theta_k}/2)|$ is a monotonically decreasing function of the vector dot product $\hat{e}_k\cdot e_k$.
\end{lemma}
\begin{proof} $|\sin{(\theta_k/2})|=\sqrt{\frac{1-\cos{(\theta_k)}}{2}}=\sqrt{\frac{1-(\hat{e}_k\cdot e_k)}{2}}$
\end{proof}

\begin{lemma}
$|\sin{(\theta_k})|$ is a monotonically decreasing function of $(\hat{e}_k\cdot e_k)^2$.
\end{lemma}

\begin{proof} $|\sin{(\theta_k})|=\sqrt{1-\cos^2(\theta_k)}=\sqrt{1-(\hat{e}_k\cdot e_k)^2}$
\end{proof}
\begin{theorem}
$L((e_x,e_y,e_z),(\hat{e}_x,\hat{e}_y,\hat{e}_z))=L((e_x,e_y,e_z),(-\hat{e}_x,\hat{e}_y,\hat{e}_z))$, i.e., the training objective (Eq.\ref{eq:loss}) respects the inversion symmetry of pseudovector $\hat{e}_x$.
\end{theorem}
\begin{proof}
\begin{align*}
\mathcal{L}((e_x,e_y,e_z),&(\hat{e}_x,\hat{e}_y,\hat{e}_z)) =\beta_x|\sin{\theta_x}| + \beta _y|\sin{\frac{\theta_y}{2}}| + \beta_z|\sin{\frac{\theta_z}{2}}|\\
&=\beta_x\sqrt{1-(\hat{e}_x\cdot e_x)^2} + \beta_y\sqrt{\frac{1-\hat{e}_y\cdot e_y}{2}} +\beta_z \sqrt{\frac{1-\hat{e}_z\cdot e_z}{2}}\\
&=\beta_x\sqrt{1-(-\hat{e}_x\cdot e_x)^2} + \beta_y\sqrt{\frac{1-\hat{e}_y\cdot e_y}{2}} +\beta_z \sqrt{\frac{1-\hat{e}_z\cdot e_z}{2}}\\
&=\mathcal{L}((e_x,e_y,e_z),(-\hat{e}_x,\hat{e}_y,\hat{e}_z))
\end{align*}
\end{proof}

We observed experimentally that our choice of loss function produced more robust results than dot product, geodesic loss, and other commonly used variants (Table~\ref{tab:ablation}).

\section{Pose Annotations}
\label{sec:annotations}

\subsubsection{Manual.} We manually annotated all poses in Research-Fetal, dHCP, and Clin\-ical-Young using ground-truth brain/eyes segmentations. We define the L$\rightarrow$R direction as pointing from the CoM of the left eye to that of the right. We define the P$\rightarrow$A direction as pointing from the brain CoM to the midpoint between the eyes, rotated 25$^\circ$ about the L$\rightarrow$R axis. All annotated poses were visually verified by a radiologist.

\subsubsection{Semi-automated.} To feasibly annotate poses in Navigators, a much larger dataset (1210 total volumes), we utilized algorithmically generated poses to assist with annotation. First, we algorithmically obtained $P_k$ values for all slices, averaged across 10 runs of optimization-based slice-to-volume registration~\cite{xu2023} (recall from Appendix~\ref{sec:interleaved} that slices are prescribed according to $\tilde{P}_k=T_kP_k$). Second, we kept “high-certainty” slices with rotation and translation errors under 8$^\circ$ and 8mm, respectively, relative to the average across all runs. Third, we adjusted each slice pose relative to the initial pose in each time-series, which was manually annotated. Fourth, since all $\tilde{P}_k$ values were available to us as part of the acquisition parameters, we computed $T_k=\tilde{P}_kP_k^{-1}$ as the GT pose for every navigator volume preceding the $k$th slice in the stack. Lastly, we manually corrected the poses in a subset of navigator volumes.

\section{Training Details}
\label{sec:training}

\subsubsection{Segmentation Network.} We train the segmentation U-Net for 1000 epochs using Adam optimization~\cite{kingma2015} with batch size 4 and learning rate $10^{-4}$. We use a weighted sum of the cross-entropy and Dice losses~\cite{milletari2016}, with weights of 1 and 0.5, respectively. We re-weight both loss terms by 8, 2, and 1 for the brain, eyes, and background classes, respectively. During training and inference, we resample all input volumes to 3mm isotropic voxels before padding to $128^3$ voxels.

We spatially augment training volumes with random rotations, translations, and scaling uniformly sampled from SO(3), $[-30,30]^3$mm, and $[0.5,1.3]$, respectively. We also simulate low resolution in 75\% of training volumes, with isotropic voxel size sampled uniformly from $[3,8]$mm. We add Gaussian noise with $\sigma$ sampled uniformly from $[0,0.03]$, random gamma correction with $\log{\gamma}$ sampled uniformly from $[-0.8,0]$, and bias field artifacts. For the bias field simulation, we use existing software~\cite{perezgarcia2021} to model the artifact as a linear combination of polynomial basis functions of order 3, with coefficients sampled uniformly from $[0,0.5]$. For spin history artifact simulation, we sample $n_{\text{slice}}$ uniformly from the unit sphere and $\sigma$ uniformly from $[1.5,2.3]$mm. We sample $c_{\text{slice}}$ such that the likelihood of a point is inversely proportional to its distance from the brain boundary. In navigator volumes, when the artifact intersects regions near the boundary of the brain, there is greater uncertainty in disambiguating the brain from external uterine tissue. Thus, we provide the network with more examples of these challenging cases during training.

\subsubsection{E(3)-CNN.} We train the E(3)-CNN regressor for 2500 epochs using stochastic gradient descent with batch size 1, learning rate $10^{-2}$, weight decay $3\times 10^{-5}$, and momentum 0.99. We spatially augment training volumes with random rotations uniformly sampled from SO(3). We simulate low resolution in 90\% of training volumes, with isotropic voxel size sampled uniformly from $[3,7.5]$mm. We use gamma augmentations with $\log{\gamma}$ sampled uniformly from $[-2.0,0.1]$ and the same bias field artifact as the segmentation network. For spin history artifact simulation, we sample $n_{\text{slice}}$ uniformly from the unit sphere and $\sigma$ uniformly from $[2.3,4.6]$mm. We sample $c_{\text{slice}}$ uniformly from voxels in the GT brain segmentation.

\section{Datasets}
\label{sec:datasets}

In this section, we provide additional information on all datasets.

\subsubsection{Research-Fetal.} This dataset consists of whole-uterus EPI volumes in 153 pregnant volunteers (3T, Siemens, 3mm isotropic resolution, TR=2.9-4s, TE=\\32-47 ms, flip angle=90$^\circ$). Fetus GA ranges from 18 to 38 weeks, with mean GA $29.0\pm 5.0$ weeks. The average FOV size is $300\!\times\!300\!\times\!156$mm.

\subsubsection{dHCP.} This dataset consists of fetal brain EPI volumes in 245 pregnant volunteers (3T, Philips Achieva, 2.2mm isotropic resolution, TR=2.2s, TE=60ms, flip angle=90$^\circ$)~\cite{karolis2025}. Fetus GA ranges from 20 to 38 weeks, with mean GA $28.9\pm 3.8$ weeks. The average FOV size is $143\!\times\!151\!\times\!108$mm. Volumes are cropped around the fetal head and are not whole-uterus volumes. Our manually annotated brain/eyes segmentations and poses for the dHCP dataset are publicly available on the project website for full transparency and future use.

\subsubsection{Clinical-Young.} This dataset consists of routine clinical whole-uterus EPI volumes in 60 pregnant patients (3T, Siemens, 1.76-3.5mm voxels, TR=2.2-5s, TE=37ms, flip angle=90$^\circ$). There are 10 subjects for each GA in the range of 18 to 23 weeks. The average FOV size is $301 \!\times \!301 \!\times\! 129$mm. Additionally, we note that while Clinical-Young is not representative of our target application of automated slice prescription, it is representative of routine clinical data acquired for retrospective head pose tracking, which promises to enable early disease detection. Specifically, there is well-supported evidence that fetal motion patterns correlate with neurodevelopmental disorders and intrauterine complications~\cite{deVries08,ayala24}.

\subsubsection{Navigators.} This dataset consists of 47 stacks of 2D HASTE slices (3T, Sie\-mens, TR=3-3.5s, TE=100ms, $1.25\!\times\!1.25\!\times\!3$mm, FA=90$^\circ$) interleaved with 3D EPI navigator volumes (3T, Siemens, TR=25-46ms, 4-6mm, TE=12-22ms, flip angle=5$^\circ$) in 9 pregnant volunteers. Fetus GA ranges from 26 to 36 weeks, with mean GA $30.6\pm 2.9$ weeks. The average FOV size is $327\!\times\!327\!\times\!125$mm. This dataset is most representative of automated slice prescription, and navigator volumes contain real spin history artifacts.

\section{Baseline Methods}
\label{sec:baselines}

We describe our implementations of the baseline methods and the modifications we make in order to optimize performance of baseline algorithms for our experiments. All methods that rely on network training were trained on a Nvidia RTX 6000 Ada GPU with batch size of either 4 or 5, unless otherwise stated.

\subsubsection{FireANTs~\cite{jena2024}.} We initialize pose estimates with moments-of-inertia matching, followed by multi-resolution adaptive gradient optimization with learning rate $3\!\times\!10^{-4}$ at downsampling factors of 4, 2, and 1 with 200, 100, and 50 iterations respectively. We find that the initialization step helps boost performance particularly in the case of large rotational misalignment between the input and template volumes.

\subsubsection{EquiTrack~\cite{billot2024}.} This method first uses a ``denoiser" CNN to remove intensity differences between input and template volumes, followed by a SE(3)-equivariant network that learns matching landmarks for rigid registration. Here, we slightly modify EquiTrack to guarantee fair comparison with our method on Navigators. First, to account for spin history artifacts, we train the denoiser network using our data augmentation strategy, which simulates these artifacts. Second, to further boost robustness to disruptive intensity perturbations (e.g., spin history artifacts), we use RANSAC~\cite{fischler1981} to compute the optimal rigid transform from matching landmarks. We train the denoiser and equivariant network for 550 and 2500 epochs with batch sizes of 4 and 1, respectively. Although we observed no statistically significant differences in performance statistics in Research-Fetal and dHCP, our modifications led to significant improvements of 5.1$^\circ$ and 10.2$^\circ$ in rotation error in Clinical-Young and Navigators, respectively.

\subsubsection{Fetal-Align~\cite{hoffmann2021}.} The original implementation uses standard image processing techniques (i.e., detection of maximally stable extremal regions~\cite{matas2004}) conditioned on GA, rather than deep learning, to estimate brain and eye masks in fetal brain volumes. We find that this approach to segmentation becomes unstable when applied to our data. For fair evaluation of landmark-based pose estimation, we use the brain and eye masks predicted by our segmentation U-Net trained on simulated spin history artifacts. When eye detection fails (a common occurrence when the artifact obstructs the eyes in navigator volumes), we return the identity transformation as an estimate of rotation. Our substitution of hand-crafted image features for deep learning-based segmentation produced improvements in rotation error of 68.9$^\circ$ and 22.4$^\circ$ in Research-Fetal and dHCP, respectively; the improvements in clinical volumes were larger.

\subsubsection{3DPose-Net~\cite{salehi2019}.} For our training experiments on Research-Fetal, we use the same augmentations as that of E(3)-Pose. For our training experiments on dHCP, we tune the data augmentation parameters such that $\log{\gamma}$ is sampled uniformly from [-3,-1] and voxel size is sampled uniformly from [3,9]mm. We train the network for 8000 epochs.

\subsubsection{6DRep~\cite{faghihpirayesh2023}.} We use the same training augmentations as 3DPose-Net and train the network for 7500 epochs.

\subsubsection{RbR~\cite{gopinath2024}.} We find that this method provides more stable and accurate pose estimates on fetal brain volumes that are preprocessed with the same steps as our E(3)-CNN, i.e., inputs that are center-of-mass aligned and scaled based on the size of the brain. Since center-of-mass alignment of input volumes eliminates the need for translation estimation, we keep the translation estimation of E(3)-Pose and predict only rotation with the output of RbR. We use the same training augmentations as 3DPose-Net. We use Adam optimization with learning rate $10^{-4}$ to train the network for 4000 epochs.

\section{Additional Results}
\label{sec:additional_results}

\subsubsection{Evaluation of Translation Estimation.} We separately evaluate the accuracy of translation estimation with our segmentation network by providing the translation error $||\hat{t}-t||_2$. We compare E(3)-Pose to baseline methods that estimate translation differently from E(3)-Pose, i.e., FireANTs and EquiTrack. Tables~\ref{tab:translation} and \ref{tab:navigators_baselines_trans_error} show that E(3)-Pose consistently outperforms EquiTrack across all datasets, and outperforms FireANTs in test volumes from Research-Fetal and Navigators test volumes. As detailed in Section~\ref{sec:results_research} in the main paper, high quality imaging (which is not representative of clinical applications in fetal MRI) facilitates the accurate rigid registration of FireANTs in dHCP test volumes. Furthermore, Table~\ref{tab:baselines} in the main paper reports that our method outperforms FireANTs in our evaluation of rotation estimation and full pose estimation in Clinical-Young. This result underscores how full 6-DoF pose estimation in fetal MRI is more challenging and useful compared to CoM alignment only. In Navigators, the presence of spin history artifacts breaks the assumption of similarity of the intensity distributions between the input and template volumes. Compared to the template-based alignment strategy, template-free pose regression improves robustness to highly disruptive artifacts.

\begin{table*}[t]
\centering
\setlength\tabcolsep{2pt}
\fontsize{7}{8.5}\selectfont
\caption{\textbf{Performance statistics for translation estimation.} Mean $\pm$ standard deviation for translation error (mm) are reported for E(3)-Pose and baseline methods that estimate translation differently from E(3)-Pose. Best score is in bold. * indicates statistical significance compared to E(3)-Pose ($p\!<\!0.05$, Bonferroni correction). We evaluate the instance of E(3)-Pose trained on Research-Fetal since translation estimation relies only on brain segmentation. See Section~\ref{sec:significance} for  details on statistical significance testing and Table~\ref{tab:navigators_baselines_trans_error} for subject-level statistics in Navigators.}
\resizebox{\textwidth}{!}{%
\begin{tabular}{lcrclcrcl}
\multicolumn{1}{c}{} & \textbf{Research-Fetal test} & \multicolumn{3}{c}{\textbf{dHCP test}} & \textbf{Clinical-Young} & \multicolumn{3}{c}{\textbf{Navigators}} \\
\hline 
FireANTs~\cite{jena2024} & \hspace{0.1cm} 2.0 $\pm$ 2.3* & \textbf{0.5} & $\mathbf{\pm}$ & \textbf{1.6*} & \textbf{0.6} $\mathbf{\pm}$ \textbf{0.5*} & 4.2 & $\pm$ & 3.1 \\
EquiTrack~\cite{billot2024} & \hspace{0.1cm} 1.3 $\pm$ 0.9* & 1.1 & $\pm$ & 0.7* & 1.6 $\pm$ 1.6* & 4.4 & $\pm$ & 3.2 \\
E(3)-Pose (ours) & \textbf{0.9} $\mathbf{\pm}$ \textbf{0.8} & 0.8 & $\pm$ & 0.8 & 0.9 $\pm$ 1.1 \hspace{0.1cm} & \textbf{3.8} & $\mathbf{\pm}$ & \textbf{2.8}\\
\hline
\end{tabular}
}
\label{tab:translation}
\end{table*}

\begin{figure}
\centering
\centerline{\includegraphics[width=0.9\textwidth]{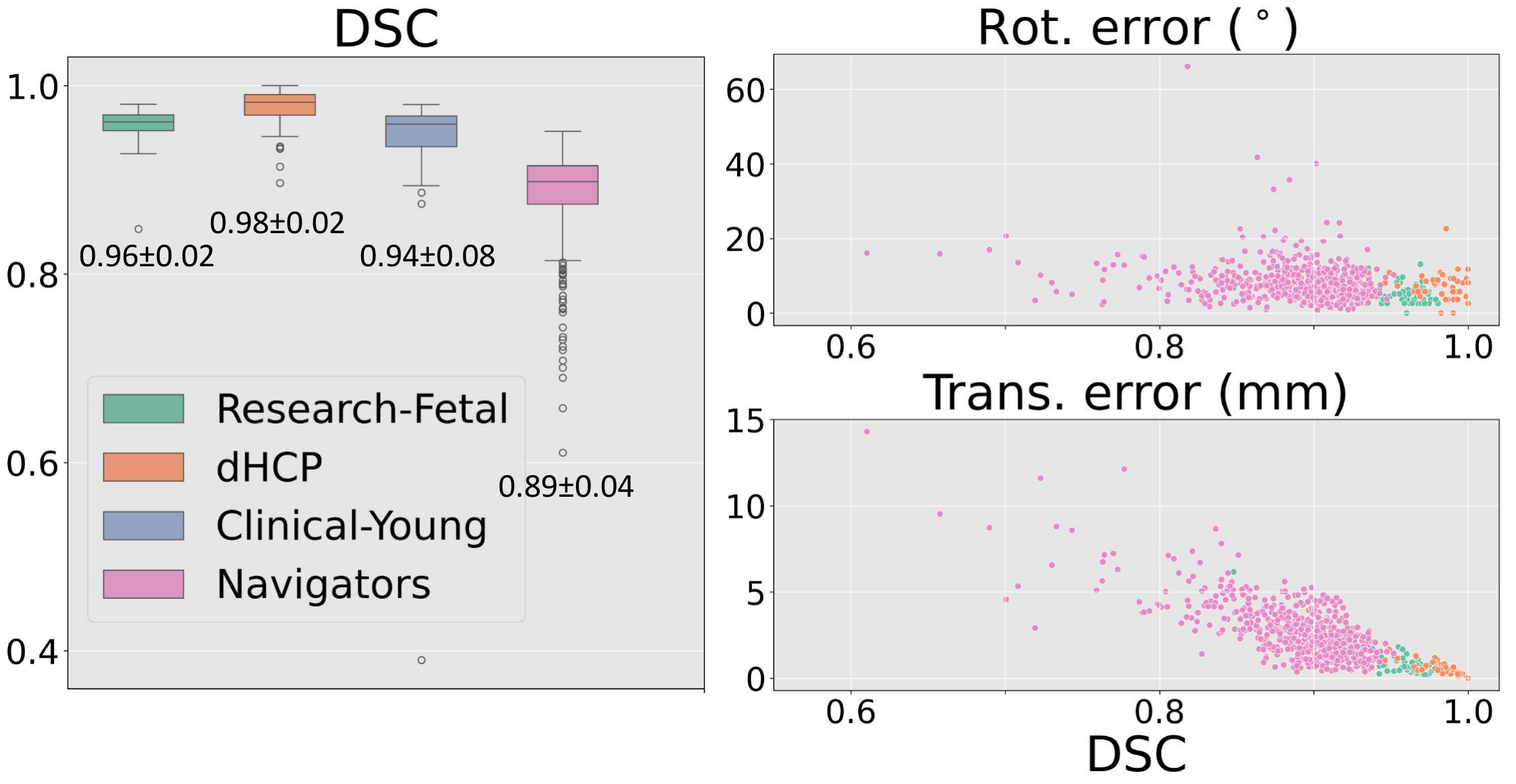}}
\caption{\textbf{Segmentation study.} Our segmentation network (trained on Research-Fetal) predicts accurate brain masks on all datasets, including challenging, out-of-distribution navigator volumes (\textit{left}). Rotation estimation with the E(3)-CNN remains robust to larger segmentation errors observed in Navigators (\textit{top right}), while translation estimation displays sensitivity to error (\textit{bottom right}).}
\label{fig:segmentation}
\end{figure}

\subsubsection{Evaluation of Segmentation.} We quantify the performance of our segmentation network by measuring the overlap between predicted and ground-truth brain masks with the Dice-Sorensen Coefficient (DSC)~\cite{zou2004}. Our segmentation network yields mean DSC of 0.96, 0.98, 0.94, and 0.89 on Research-Fetal, dHCP, Clinical-Young, and Navigators, respectively (Fig.~\ref{fig:segmentation}, left). While brain segmentation is robust across datasets, larger errors tend to occur in Navigators, which pose a challenging domain gap from the training data. We further observe that rotation estimation using our E(3)-CNN remains robust to segmentation errors, particularly larger errors in Navigators (Fig.~\ref{fig:segmentation}, top right). This result highlights the advantages of a neural architecture constructed with strong inductive biases, coupled with a cropping margin large enough for the input volume to capture relevant anatomy regardless of segmentation errors. However, translation estimation is sensitive to segmentation error, since it is directly computed from the predicted mask (Fig.~\ref{fig:segmentation}, bottom right).

\subsubsection{Statistical Significance Testing.}
\label{sec:significance}

\begin{table*}
\centering
\caption{\textbf{Subject-level baseline comparisons for rotation error ($^\circ$) in Navigators.}  Mean $\pm$ 95\% confidence interval statistics are reported. Subjects are denoted by gestational age (GA, weeks). We separately report the performance statistics for FireANTs and Fetal-Align, which do not rely on training (beyond segmentation). Best score by training setting is in bold. Best overall score is italicized. E(3)-Pose outperforms the baseline methods on most subjects in Navigators, across training environments.}
\setlength\tabcolsep{1pt}
\fontsize{7}{7.5}\selectfont
\resizebox{\textwidth}{!}{%
\begin{tabular}{lrclrclrclrclrclrclrclrclrcl}
\multicolumn{1}{c}{Subject (GA) } & \multicolumn{3}{c}{S1 (28w)} & \multicolumn{3}{c}{S2 (28w)} & \multicolumn{3}{c}{S3 (32w)} & \multicolumn{3}{c}{S4 (34w)} & \multicolumn{3}{c}{S5 (36w)} & \multicolumn{3}{c}{S6 (28w)} & \multicolumn{3}{c}{S7 (26w)} & \multicolumn{3}{c}{S8 (31w)} & \multicolumn{3}{c}{S9 (28w)}\\
\midrule
\multicolumn{17}{l}{\textit{No training}} \\
\midrule
FireANTs~\cite{jena2024} &  65.7 & $\pm$ & 8.2 & \textbf{62.8} & $\pm$ & \textbf{8.6} & \textbf{49.4} & $\pm$ & \textbf{12.9} & \textbf{10.0} & $\pm$ & \textbf{1.2} & \textbf{53.5} & $\pm$ & \textbf{8.1} & 75.8 & $\pm$ & 14.3 & 28.7 & $\pm$ & 6.4 & \textbf{8.6} & $\pm$ & \textbf{1.7} & \textbf{24.0} & $\pm$ & \textbf{8.6} \\
Fetal-Align~\cite{hoffmann2021} & \textbf{13.2} & $\pm$ & \textbf{4.1} & 90.3 & $\pm$ & 5.3 & 52.3 & $\pm$ & 10.5 & 35.7 & $\pm$ & 8.6 & 115.3 & $\pm$ & 8.4 & \textbf{62.3} & $\pm$ & \textbf{11.2} & \textbf{21.3} & $\pm$ & \textbf{4.2} & 10.7 & $\pm$ & 3.9 & 29.4 & $\pm$ & 9.3\\
\midrule
\multicolumn{17}{l}{\textit{Trained on Research-Fetal}} \\
\midrule
EquiTrack~\cite{billot2024} & 40.1 & $\pm$ & 6.0 & 63.9 & $\pm$ & 6.6 & 37.7 & $\pm$ & 8.2 & 45.5 & $\pm$ & 9.8 & 45.3 & $\pm$ & 6.6 & 97.5 & $\pm$ & 12.6 & 24.2 & $\pm$ & 4.3 & 28.5 & $\pm$ & 8.2 & 21.3 & $\pm$ & 8.5\\
3DPose-Net~\cite{salehi2019} & 54.7 & $\pm$ & 6.0 & 113.4 & $\pm$ & 5.6  & 98.6 & $\pm$ & 8.7 & 56.4 & $\pm$ & 6.7 & 30.2 & $\pm$ & 2.8 & 101.5 & $\pm$ & 9.3 & 65.9 & $\pm$ & 7.6 & 35.2 & $\pm$ & 6.1 & 37.8 & $\pm$ & 4.5\\
6DRep~\cite{faghihpirayesh2023} & 23.1 & $\pm$ & 2.5 & 95.0 & $\pm$ & 7.0 & 24.4 & $\pm$ & 3.9 & 36.2 & $\pm$ & 7.8 & 11.7 & $\pm$ & 1.8 & 52.6 & $\pm$ & 9.7 & 46.3 & $\pm$ & 8.7 & 18.5 & $\pm$ & 5.1 & 19.2 & $\pm$ & 1.6\\
RbR~\cite{gopinath2024} & 20.7 & $\pm$ & 2.9 & 43.3 & $\pm$ & 3.8 & 20.6 & $\pm$ & 3.8 & 10.7 & $\pm$ & 1.2 & \textbf{10.4} & $\mathbf{\pm}$ & \textbf{1.2} & 51.8 & $\pm$ & 10.1  & 26.0 & $\pm$ & 3.0 & 11.0 & $\pm$ & 0.9 & 9.3 & $\pm$ & 0.8\\
E(3)-Pose (ours) & \textit{\textbf{6.4}} & $\mathbf{\pm}$ & \textit{\textbf{0.5}} & \textit{\textbf{13.6}} & $\mathbf{\pm}$ & \textit{\textbf{1.5}} & \textit{\textbf{10.6}} & $\mathbf{\pm}$ & \textit{\textbf{3.4}} & \textbf{6.6} & $\mathbf{\pm}$ & \textbf{0.6}  & 10.7 & $\pm$ & 0.8 & \textit{\textbf{13.4}} & $\mathbf{\pm}$ & \textit{\textbf{1.9}} & \textit{\textbf{9.2}} & $\mathbf{\pm}$ & \textit{\textbf{0.8}} & \textbf{8.4} & $\mathbf{\pm}$ & \textbf{1.0}& \textit{\textbf{7.4}} & $\mathbf{\pm}$ & \textit{\textbf{0.6}}\\
\midrule
\multicolumn{17}{l}{\textit{Trained on dHCP}} \\
\midrule
EquiTrack~\cite{billot2024} & 32.9 & $\pm$ & 5.8 & 66.3 & $\pm$ & 6.8 & 65.2 & $\pm$ & 12.2 & 75.5 & $\pm$ & 9.9 & 76.7 & $\pm$ & 8.5 & 109.4 & $\pm$ & 10.4 & 18.8 & $\pm$ & 3.8 & 68.5 & $\pm$ & 13.4 & 36.4 & $\pm$ & 10.9\\
3DPose-Net~\cite{salehi2019} & 138.8 & $\pm$ & 3.1 & 100.2 & $\pm$ & 3.5 & 139.8 & $\pm$ & 4.7 & 115.7 & $\pm$ & 4.7 & 16.8 & $\pm$ & 1.7 & 69.4 & $\pm$ & 4.7 & 79.6 & $\pm$ & 3.0 & 54.1 & $\pm$ & 3.7 & 102.9 & $\pm$ & 2.9\\
6DRep~\cite{faghihpirayesh2023} & 55.2 & $\pm$ & 4.2 & 153.1 & $\pm$ & 4.2 & 72.6 & $\pm$ & 2.9 & 82.2 & $\pm$ & 12.7 & 17.0 & $\pm$ & 1.1 & 38.6 & $\pm$ & 3.3 & 137.8 & $\pm$ & 3.6 & 79.5 & $\pm$ & 4.2 & 115.4 & $\pm$ & 7.7 \\
RbR~\cite{gopinath2024} & 45.3 & $\pm$ & 3.2 & 121.0 & $\pm$ & 4.1 & 100.8 & $\pm$ & 5.4 & 41.1 & $\pm$ & 6.4 & 23.1 & $\pm$ & 4.1 & 118.6 & $\pm$ & 8.0  & 145.6 & $\pm$ & 4.5 & 86.3 & $\pm$ & 8.0 & 37.1 & $\pm$ & 4.9 \\
E(3)-Pose (ours) & \textbf{12.7} & $\mathbf{\pm}$ & \textbf{1.6} & \textbf{24.4} & $\mathbf{\pm}$ & \textbf{2.6} & \textbf{14.4} & $\mathbf{\pm}$ & \textbf{3.3} & \textit{\textbf{6.5}} & $\mathbf{\pm}$ & \textit{\textbf{0.5}} & \textit{\textbf{9.2}} & $\mathbf{\pm}$ & \textit{\textbf{0.9}} & \textbf{14.9} & $\mathbf{\pm}$ & \textbf{1.7} & \textbf{16.9} & $\mathbf{\pm}$ & \textbf{2.2} & \textit{\textbf{7.0}} & $\mathbf{\pm}$ & \textit{\textbf{0.8}} & \textbf{14.2} & $\mathbf{\pm}$ & \textbf{1.7}\\
\hline
\end{tabular}
}
\label{tab:navigators_baselines_rot_error}
\end{table*}

\begin{table*}
\centering
\caption{\textbf{Subject-level baseline comparisons for average absolute distance (AAD, mm) in Navigators.}   Mean $\pm$ 95\% confidence interval statistics are reported. Subjects are denoted by gestational age (GA, weeks). We separately report the performance statistics for FireANTs and Fetal-Align, which do not rely on training (beyond segmentation). Best score by training setting is in bold. Best overall score is italicized. E(3)-Pose outperforms the baseline methods on most subjects in Navigators, across training environments.}
\setlength\tabcolsep{1pt}
\fontsize{7}{7.5}\selectfont
\resizebox{\textwidth}{!}{%
\begin{tabular}{lrclrclrclrclrclrclrclrclrcl}
\multicolumn{1}{c}{Subject (GA) } & \multicolumn{3}{c}{S1 (28w)} & \multicolumn{3}{c}{S2 (28w)} & \multicolumn{3}{c}{S3 (32w)} & \multicolumn{3}{c}{S4 (34w)} & \multicolumn{3}{c}{S5 (36w)} & \multicolumn{3}{c}{S6 (28w)} & \multicolumn{3}{c}{S7 (26w)} & \multicolumn{3}{c}{S8 (31w)} & \multicolumn{3}{c}{S9 (28w)}\\
\midrule
\multicolumn{17}{l}{\textit{No training}} \\
\midrule
FireANTs~\cite{jena2024} &  24.4 & $\pm$ & 2.5 & \textbf{22.5} & $\pm$ & \textbf{2.6} & \textbf{22.5} & $\pm$ & \textbf{4.7} & \textbf{6.2} & $\pm$ & \textbf{0.7} & \textbf{27.5} & $\pm$ & \textbf{3.8} & 33.1 & $\pm$ & 4.8 & 12.1 & $\pm$ & 2.1 & \textbf{5.8} & $\pm$ & \textbf{0.9} & \textbf{11.2} & $\pm$ & \textbf{3.0} \\
Fetal-Align~\cite{hoffmann2021} & \textbf{7.4} & $\pm$ & \textbf{1.4} & 36.2 & $\pm$ & 2.0 & 27.4 & $\pm$ & 4.8 & 18.8 & $\pm$ & 3.8 & 49.5 & $\pm$ & 3.3 & \textbf{29.0} & $\pm$ & \textbf{4.5} & \textbf{9.5} & $\pm$ & \textbf{1.6} & 6.0 & $\pm$ & 1.6 & 13.4 & $\pm$ & 3.8\\
\midrule
\multicolumn{17}{l}{\textit{Trained on Research-Fetal}} \\
\midrule
EquiTrack~\cite{billot2024} & 17.5 & $\pm$ & 2.1 & 25.7 & $\pm$ & 2.2 & 19.9 & $\pm$ & 3.2 & 21.9 & $\pm$ & 3.8 & 23.7 & $\pm$ & 2.8 & 41.4 & $\pm$ & 4.1 & 11.4 & $\pm$ & 1.5 & 13.0 & $\pm$ & 2.9 & 9.3 & $\pm$ & 2.9\\
3DPose-Net~\cite{salehi2019} & 23.1 & $\pm$ & 2.1 & 42.1 & $\pm$ & 1.6  & 43.2 & $\pm$ & 2.6 & 29.7 & $\pm$ & 2.6 & 17.5 & $\pm$ & 1.4 & 43.1 & $\pm$ & 2.7 & 24.3 & $\pm$ & 2.3 & 16.3 & $\pm$ & 2.2 & 17.2 & $\pm$ & 1.9\\
6DRep~\cite{faghihpirayesh2023} & 11.4 & $\pm$ & 1.1 & 36.5 & $\pm$ & 2.2 & 14.2 & $\pm$ & 2.0 & 18.7 & $\pm$ & 3.2 & 7.9 & $\pm$ & 0.9 & 25.9 & $\pm$ & 3.5 & 16.5 & $\pm$ & 2.5 & 9.0 & $\pm$ & 2.0 & 9.1 & $\pm$ & 0.7\\
RbR~\cite{gopinath2024} & 10.4 & $\pm$ & 1.1 & 19.4 & $\pm$ & 1.4 & 12.2 & $\pm$ & 1.8 & 7.4 & $\pm$ & 0.6 & \textbf{7.3} & $\mathbf{\pm}$ & \textbf{0.7} & 25.8 & $\pm$ & 4.1  & 11.0 & $\pm$ & 1.1 & 6.3 & $\pm$ & 0.5 & 5.0 & $\pm$ & 0.4\\
E(3)-Pose (ours) & \textit{\textbf{5.0}} & $\mathbf{\pm}$ & \textit{\textbf{0.3}} & \textit{\textbf{7.6}} & $\mathbf{\pm}$ & \textit{\textbf{0.7}} & \textit{\textbf{7.5}} & $\mathbf{\pm}$ & \textbf{1.7} & \textbf{5.5} & $\mathbf{\pm}$ & \textbf{0.2}  & 7.4 & $\pm$ & 0.5 & \textit{\textbf{10.1}} & $\mathbf{\pm}$ & \textit{\textbf{1.2}} & \textit{\textbf{4.8}} & $\mathbf{\pm}$ & \textit{\textbf{0.3}} & \textbf{5.1} & $\mathbf{\pm}$ & \textbf{0.5}& \textit{\textbf{4.1}} & $\mathbf{\pm}$ & \textit{\textbf{0.3}}\\
\midrule
\multicolumn{17}{l}{\textit{Trained on dHCP}} \\
\midrule
EquiTrack~\cite{billot2024} & 14.8 & $\pm$ & 2.2 & 26.1 & $\pm$ & 2.3 & 29.8 & $\pm$ & 4.7 & 36.2 & $\pm$ & 4.1 & 36.8 & $\pm$ & 3.4 & 45.9 & $\pm$ & 2.7 & 8.3 & $\pm$ & 1.3 & 25.2 & $\pm$ & 4.4 & 14.7 & $\pm$ & 3.7\\
3DPose-Net~\cite{salehi2019} & 49.5 & $\pm$ & 0.6 & 38.2 & $\pm$ & 1.1 & 60.1 & $\pm$ & 1.0 & 54.0 & $\pm$ & 1.5 & 10.5 & $\pm$ & 0.9 & 33.8 & $\pm$ & 1.8 & 31.2 & $\pm$ & 1.0 & 26.5 & $\pm$ & 1.5 & 40.9 & $\pm$ & 0.9\\
6DRep~\cite{faghihpirayesh2023} & 23.9 & $\pm$ & 1.6 & 49.8 & $\pm$ & 1.0 & 36.0 & $\pm$ & 1.3 & 35.2 & $\pm$ & 4.7 & 10.8 & $\pm$ & 10.6 & 20.4 & $\pm$ & 1.5 & 44.5 & $\pm$ & 0.7 & 36.8 & $\pm$ & 1.7 & 45.0 & $\pm$ & 2.0 \\
RbR~\cite{gopinath2024} & 20.4 & $\pm$ & 1.2 & 43.5 & $\pm$ & 0.9 & 48.3 & $\pm$ & 1.8 & 21.7 & $\pm$ & 2.8 & 13.2 & $\pm$ & 1.8 & 48.5 & $\pm$ & 2.4 & 45.4 & $\pm$ & 0.8 & 34.7 & $\pm$ & 2.7 & 17.1 & $\pm$ & 1.8 \\
E(3)-Pose (ours) & \textbf{7.3} & $\mathbf{\pm}$ & \textbf{0.7} & \textbf{11.9} & $\mathbf{\pm}$ & \textbf{1.0} & \textbf{9.2} & $\mathbf{\pm}$ & \textbf{1.6} & \textit{\textbf{5.4}} & $\mathbf{\pm}$ & \textit{\textbf{0.3}} & \textit{\textbf{6.7}} & $\mathbf{\pm}$ & \textit{\textbf{0.6}} & \textbf{10.6} & $\mathbf{\pm}$ & \textbf{1.1} & \textbf{7.9} & $\mathbf{\pm}$ & \textbf{0.9} & \textit{\textbf{4.5}} & $\mathbf{\pm}$ & \textit{\textbf{0.4}} & \textbf{7.2} & $\mathbf{\pm}$ & \textbf{0.8}\\
\hline
\end{tabular}}
\label{tab:navigators_baselines_aad}
\end{table*}

\begin{table*}
\centering
\caption{\textbf{Subject-level baseline comparisons for translation error (mm) in Navigators.}  Mean $\pm$ 95\% confidence interval statistics are reported for E(3)-Pose and baseline methods that estimate translation differently from E(3)-Pose. We evaluate the instance of E(3)-Pose trained on Research-Fetal since translation estimation relies only on brain segmentation. Subjects are denoted by gestational age (GA, weeks). Best score is in bold. E(3)-Pose outperforms the baseline methods on most subjects in Navigators.}
\setlength\tabcolsep{1pt}
\fontsize{7}{7.5}\selectfont
\resizebox{\textwidth}{!}{%
\begin{tabular}{lrclrclrclrclrclrclrclrclrcl}
\multicolumn{1}{c}{Subject (GA) } & \multicolumn{3}{c}{S1 (28w)} & \multicolumn{3}{c}{S2 (28w)} & \multicolumn{3}{c}{S3 (32w)} & \multicolumn{3}{c}{S4 (34w)} & \multicolumn{3}{c}{S5 (36w)} & \multicolumn{3}{c}{S6 (28w)} & \multicolumn{3}{c}{S7 (26w)} & \multicolumn{3}{c}{S8 (31w)} & \multicolumn{3}{c}{S9 (28w)}\\
\hline 
FireANTs~\cite{jena2024} &  \textbf{3.6} & $\mathbf{\pm}$ & \textbf{0.3} & 4.6 & $\pm$ & 0.4 & 5.8 & $\pm$ & 1.1 & \textbf{2.7} & $\mathbf{\pm}$ & \textbf{0.3} & 4.6 & $\pm$ & 0.4 & 8.1 & $\pm$ & 1.0 & 3.0 & $\pm$ & 0.2 & 3.4 & $\pm$ & 0.3 & 3.5 & $\pm$ & 0.4 \\
EquiTrack~\cite{billot2024} & 4.0 & $\pm$ & 0.3 & 5.6 & $\pm$ & 0.5 & 5.1 & $\pm$ & 1.0 & 3.7 & $\pm$ & 0.5 & 3.9 & $\pm$ & 0.4 & 7.5 & $\pm$ & 1.0 & 4.5 & $\pm$ & 0.3 & 3.1 & $\pm$ & 0.3 & 2.1 & $\pm$ & 0.3\\
E(3)-Pose (ours) & 4.0 & $\pm$ & 0.3 & \textbf{4.2} & $\mathbf{\pm}$ & \textbf{0.4} & \textbf{5.0} & $\mathbf{\pm}$ & \textbf{1.0} & 3.9 & $\pm$ & 0.2  & \textbf{3.6} & $\mathbf{\pm}$ & \textbf{0.4} & \textbf{7.4} & $\mathbf{\pm}$ & \textbf{0.9} & \textbf{2.7} & $\mathbf{\pm}$ & \textbf{0.2} & \textbf{2.6} & $\mathbf{\pm}$ & \textbf{0.2}& \textbf{2.0} & $\mathbf{\pm}$ & \textbf{0.3}\\
\hline
\end{tabular}}
\label{tab:navigators_baselines_trans_error}
\end{table*}

\begin{table*}
\centering
\caption{\textbf{Subject-level ablation study for rotation error ($^\circ$) in Navigators.}  Mean $\pm$ 95\% confidence interval statistics are reported. Subjects are denoted by gestational age (GA, weeks). Best score by training setting is in bold.}
\setlength\tabcolsep{1pt}
\fontsize{7}{7.5}\selectfont
\resizebox{\textwidth}{!}{%
\begin{tabular}{lrclrclrclrclrclrclrclrclrcl}
\multicolumn{1}{c}{Subject (GA) } & \multicolumn{3}{c}{S1 (28w)} & \multicolumn{3}{c}{S2 (28w)} & \multicolumn{3}{c}{S3 (32w)} & \multicolumn{3}{c}{S4 (34w)} & \multicolumn{3}{c}{S5 (36w)} & \multicolumn{3}{c}{S6 (28w)} & \multicolumn{3}{c}{S7 (26w)} & \multicolumn{3}{c}{S8 (31w)} & \multicolumn{3}{c}{S9 (28w)}\\
\midrule
\multicolumn{17}{l}{\textit{Trained on Research-Fetal}} \\
\midrule
E(3)-Pose (ours) & \textbf{6.4} & $\mathbf{\pm}$ & \textbf{0.5} & \textbf{13.6} & $\mathbf{\pm}$ & \textbf{1.5} & \textbf{10.6} & $\mathbf{\pm}$ & \textbf{3.4} & 6.6 & $\pm$ & 0.6  & 10.7 & $\pm$ & 0.8 & \textbf{13.4} & $\mathbf{\pm}$ & \textbf{1.9} & 9.2 & $\pm$ & 0.8 & 8.4 & $\pm$ & 1.0& 7.4 & $\pm$ & 0.6\\
Standard CNN & 10.8 & $\pm$ & 0.4 & 32.4 & $\pm$ & 4.0 & 19.8 & $\pm$ & 4.1 & 12.7 & $\pm$ & 2.3 & 10.2 & $\pm$ & 0.9 & 22.5 & $\pm$ & 4.3 & 29.0 & $\pm$ & 3.7 & 11.7 & $\pm$ & 1.7 & 7.5 & $\pm$ & 0.6\\
no pseudovector & 9.4 & $\pm$ & 0.9 & 15.8 & $\pm$ & 2.4 & 11.2 & $\pm$ & 3.2 & \textbf{5.8} & $\mathbf{\pm}$ & \textbf{0.5} & 12.6 & $\pm$ & 1.0 & 18.0 & $\pm$ & 4.0 & \textbf{8.7} & $\mathbf{\pm}$ & \textbf{1.0} & \textbf{6.2} & $\mathbf{\pm}$ & \textbf{0.6} & 8.6 & $\pm$ & 0.6\\
$h(R)\!=\!e_y\!\oplus\!e_z$ & 7.5 & $\pm$ & 0.6 & 19.7 & $\pm$ & 2.9 & 11.0 & $\pm$ & 3.6 & 6.5 & $\pm$ & 0.8 & \textbf{9.6} & $\mathbf{\pm}$ & \textbf{0.7} & 20.8 & $\pm$ & 2.5 & 14.0 & $\pm$ & 2.2 & 9.7 & $\pm$ & 0.8 & \textbf{7.3} & $\mathbf{\pm}$ & \textbf{0.7}\\
$|\sin{\frac{\theta_x}{2}}|$ & 7.9 & $\pm$ & 0.5 & 19.2 & $\pm$ & 2.2 & 12.8 & $\pm$ & 3.6 & 7.4 & $\pm$ & 0.9 & 11.1 & $\pm$ & 0.9 & 19.8 & $\pm$ & 3.6 & 10.9 & $\pm$ & 1.3 & 10.5 & $\pm$ & 0.8 & 8.7 & $\pm$ & 0.6\\
geodesic loss & 7.2 & $\pm$ & 0.5 & 25.8 & $\pm$ & 4.6 & 12.6 & $\pm$ & 3.4 & 6.4 & $\pm$ & 0.5 & 12.9 & $\mathbf{\pm}$ & 1.6 & 23.1 & $\pm$ & 5.9 & 14.1 & $\pm$ & 3.7 & \textbf{6.2} & $\mathbf{\pm}$ & \textbf{0.7} & 8.7 & $\pm$ & 1.1\\
no artifact augm. & 11.6 & $\pm$ & 1.5 & 22.7 & $\pm$ & 3.1 & 11.8 & $\pm$ & 3.6 & 15.5 & $\pm$ & 2.5 & 13.0 & $\pm$ & 1.0 & 35.8 & $\pm$ & 7.0 & 15.5 & $\pm$ & 3.2 & 11.9 & $\pm$ & 2.2 & 12.3 & $\pm$ & 4.2\\
\midrule
\multicolumn{17}{l}{\textit{Trained on dHCP}} \\
\midrule
E(3)-Pose (ours) & 12.7 & $\pm$ & 1.6 & \textbf{24.4} & $\mathbf{\pm}$ & \textbf{2.6} & \textbf{14.4} & $\mathbf{\pm}$ & \textbf{3.3} & \textbf{6.5} & $\mathbf{\pm}$ & \textbf{0.5} & 9.2 & $\pm$ & 0.9 & \textbf{14.9} & $\mathbf{\pm}$ & \textbf{1.7} & \textbf{16.9} & $\mathbf{\pm}$ & \textbf{2.2} & 7.0 & $\pm$ & 0.8 & \textbf{14.2} & $\mathbf{\pm}$ & \textbf{1.7}\\
Standard CNN & 93.9 & $\pm$ & 6.9 & 108.0 & $\pm$ & 3.7 & 147.0 & $\pm$ & 4.7 & 28.4 & $\pm$ & 3.9 & 11.7 & $\pm$ & 1.0 & 105.4 & $\pm$ & 2.9 & 111.8 & $\pm$ & 6.0 & 80.6 & $\pm$ & 12.1 & 86.3 & $\pm$ & 4.1\\
no pseudovector & 13.1 & $\pm$ & 1.9 & 45.3 & $\pm$ & 5.3 & 21.9 & $\pm$ & 4.4 & 8.9 & $\pm$ & 1.1 & 9.4 & $\pm$ & 0.9 & 32.7 & $\pm$ & 9.6 & 21.9 & $\pm$ & 3.2 & \textbf{5.9} & $\mathbf{\pm}$ & \textbf{0.7} & 23.7 & $\pm$ & 3.6\\
$h(R)\!=\!e_y\!\oplus\!e_z$ & 11.2 & $\pm$ & 1.2 & 43.7 & $\pm$ & 6.1 & 15.0 & $\pm$ & 5.0 & 8.8 & $\pm$ & 0.9 & 11.4 & $\pm$ & 1.5 & 38.1 & $\pm$ & 8.7 & 66.1 & $\pm$ & 8.3 & 10.7 & $\pm$ & 3.1 & 25.3 & $\pm$ & 5.0\\
$|\sin{\frac{\theta_x}{2}}|$ & \textbf{9.4} & $\mathbf{\pm}$ & \textbf{0.9} & 28.0 & $\pm$ & 4.3 & 16.7 & $\pm$ & 5.2 & 7.7 & $\pm$ & 0.7 & \textbf{8.7} & $\mathbf{\pm}$ & \textbf{0.8} & 29.3 & $\pm$ & 8.4 & 42.5 & $\pm$ & 7.7 & 9.5 & $\pm$ & 3.1 & 15.9 & $\pm$ & 1.5\\
geodesic loss & 18.9 & $\pm$ & 4.8 & 47.8 & $\pm$ & 7.1 & 47.5 & $\pm$ & 14.5 & 11.1 & $\pm$ & 0.8 & 13.8 & $\pm$ & 1.1 & 35.5 & $\pm$ & 8.8 & 28.6 & $\pm$ & 6.7 & 32.3 & $\pm$ & 11.1 & 22.5 & $\pm$ & 7.1\\
no artifact augm. & 40.9 & $\pm$ & 6.7 & 47.6 & $\pm$ & 5.8 & 34.3 & $\pm$ & 8.3 & 22.5 & $\pm$ & 6.1 & 16.5 & $\pm$ & 2.2 & 55.0 & $\pm$ & 9.7 & 26.9 & $\pm$ & 5.0 & 19.5 & $\pm$ & 4.4 & 28.2 & $\pm$ & 4.7\\
\hline
\end{tabular}}
\label{tab:navigators_ablations_rot_error}
\end{table*}

\begin{table*}
\centering
\caption{\textbf{Subject-level ablation study for average absolute distance (AAD, mm) in Navigators.}   Mean $\pm$ 95\% confidence interval statistics are reported. Subjects are denoted by gestational age (GA, weeks). Best score by training setting is in bold.}
\setlength\tabcolsep{1pt}
\fontsize{7}{7.5}\selectfont
\resizebox{\textwidth}{!}{%
\begin{tabular}{lrclrclrclrclrclrclrclrclrcl}
\multicolumn{1}{c}{Subject (GA) } & \multicolumn{3}{c}{S1 (28w)} & \multicolumn{3}{c}{S2 (28w)} & \multicolumn{3}{c}{S3 (32w)} & \multicolumn{3}{c}{S4 (34w)} & \multicolumn{3}{c}{S5 (36w)} & \multicolumn{3}{c}{S6 (28w)} & \multicolumn{3}{c}{S7 (26w)} & \multicolumn{3}{c}{S8 (31w)} & \multicolumn{3}{c}{S9 (28w)}\\
\midrule
\multicolumn{17}{l}{\textit{Trained on Research-Fetal}} \\
\midrule
E(3)-Pose (ours) & \textbf{5.0} & $\mathbf{\pm}$ & \textbf{0.3} & \textbf{7.6} & $\mathbf{\pm}$ & \textbf{0.7} & \textbf{7.5} & $\mathbf{\pm}$ & \textbf{1.7} & 5.5 & $\pm$ & 0.2  & 7.4 & $\pm$ & 0.5 & \textbf{10.1} & $\mathbf{\pm}$ & \textbf{1.2} & 4.8 & $\pm$ & 0.3 & 5.1 & $\pm$ & 0.5& \textbf{4.1} & $\mathbf{\pm}$ & \textbf{0.3}\\
Standard CNN & 6.3 & $\pm$ & 0.3 & 14.9 & $\pm$ & 1.5 & 12.0 & $\pm$ & 1.9 & 8.9 & $\pm$ & 1.1 & 7.6 & $\pm$ & 0.5 & 14.0 & $\pm$ & 1.9 & 11.9 & $\pm$ & 1.2 & 6.5 & $\pm$ & 0.8 & 4.3 & $\pm$ & 0.4\\
no pseudovector & 5.9 & $\pm$ & 0.4 & 8.7 & $\pm$ & 0.9 & 8.5 & $\pm$ & 1.5 & \textbf{5.3} & $\mathbf{\pm}$ & \textbf{0.2} & 8.7 & $\pm$ & 0.6 & 11.9 & $\pm$ & 1.8 & \textbf{4.7} & $\mathbf{\pm}$ & \textbf{0.4} & \textbf{4.1} & $\mathbf{\pm}$ & \textbf{0.3} & 4.8 & $\pm$ & 0.3\\
$h(R)\!=\!e_y\!\oplus\!e_z$ & 5.5 & $\pm$ & 0.3 & 10.1 & $\pm$ & 1.1 & 8.2 & $\pm$ & 1.6 & 5.7 & $\pm$ & 0.4 & \textbf{7.3} & $\mathbf{\pm}$ & \textbf{0.5} & 13.2 & $\pm$ & 1.3 & 6.5 & $\pm$ & 0.7 & 5.5 & $\pm$ & 0.4 & \textbf{4.1} & $\mathbf{\pm}$ & \textbf{0.3}\\
$|\sin{\frac{\theta_x}{2}}|$ & 5.4 & $\pm$ & 0.3 & 9.7 & $\pm$ & 0.9 & 8.4 & $\pm$ & 1.7 & 5.8 & $\pm$ & 0.5 & 7.6 & $\pm$ & 0.6 & 12.7 & $\pm$ & 1.6 & 5.5 & $\pm$ & 0.5 & 5.9 & $\pm$ & 0.4 & 4.8 & $\pm$ & 0.3\\
geodesic loss & 5.1 & $\pm$ & 0.3 & 11.7 & $\pm$ & 1.6 & 8.4 & $\pm$ & 1.6 & \textbf{5.3} & $\mathbf{\pm}$ & \textbf{0.3} & 8.5 & $\pm$ & 10.8 & 14.1 & $\pm$ & 2.4 & 6.4 & $\pm$ & 1.2 & \textbf{4.1} & $\mathbf{\pm}$ & \textbf{0.4} & 4.7 & $\pm$ & 0.5\\
no artifact augm. & 6.8 & $\pm$ & 0.6 & 11.2 & $\pm$ & 1.2 & 8.7 & $\pm$ & 1.6 & 9.5 & $\pm$ & 21.1 & 8.9 & $\pm$ & 0.6 & 18.9 & $\pm$ & 2.6  & 7.1 & $\pm$ & 1.1 & 6.6 & $\pm$ & 1.0 & 6.1 & $\pm$ & 1.6\\
\midrule
\multicolumn{17}{l}{\textit{Trained on dHCP}} \\
\midrule
E(3)-Pose (ours) & 7.3 & $\pm$ & 0.7 & \textbf{11.9} & $\mathbf{\pm}$ & \textbf{1.0} & \textbf{9.2} & $\mathbf{\pm}$ & \textbf{1.6} & \textbf{5.4} & $\mathbf{\pm}$ & \textbf{0.3} & 6.7 & $\pm$ & 0.6 & \textbf{10.6} & $\mathbf{\pm}$ & \textbf{1.1} & \textbf{7.9} & $\mathbf{\pm}$ & \textbf{0.9} & 4.5 & $\pm$ & 0.4 & \textbf{7.2} & $\mathbf{\pm}$ & \textbf{0.8}\\
Standard CNN & 35.0 & $\pm$ & 2.0 & 39.4 & $\pm$ & 0.8 & 57.0 & $\pm$ & 0.7 & 16.1 & $\pm$ & 1.9 & 8.0 & $\pm$ & 0.6 & 47.1 & $\pm$ & 1.1 & 38.6 & $\pm$ & 1.2 & 32.0 & $\pm$ & 4.0 & 39.3 & $\pm$ & 1.6\\
no pseudovector & 7.3 & $\pm$ & 0.8 & 19.9 & $\pm$ & 2.0 & 13.0 & $\pm$ & 2.0 & 6.5 & $\pm$ & 0.6 & 6.8 & $\pm$ & 0.6 & 16.9 & $\pm$ & 3.2 & 9.6 & $\pm$ & 1.2 & \textbf{4.0} & $\mathbf{\pm}$ & \textbf{0.3} & 11.5 & $\pm$ & 1.5\\
$h(R)\!=\!e_y\!\oplus\!e_z$ & 6.8 & $\pm$ & 0.5 & 18.2 & $\pm$ & 2.0 & 9.4 & $\pm$ & 2.1 & 6.4 & $\pm$ & 0.4 & 7.9 & $\pm$ & 0.9 & 20.3 & $\pm$ & 3.5 & 24.1 & $\pm$ & 2.5 & 6.0 & $\pm$ & 1.1 & 12.2 & $\pm$ & 2.1\\
$|\sin{\frac{\theta_x}{2}}|$ & \textbf{6.1} & $\mathbf{\pm}$ & \textbf{0.4} & 12.7 & $\pm$ & 1.5 & 10.1 & $\pm$ & 2.3 & 6.0 & $\pm$ & 0.3 & \textbf{6.4} & $\mathbf{\pm}$ & \textbf{0.5} & 16.0 & $\pm$ & 3.1  & 16.1 & $\pm$ & 2.4 & 5.5 & $\pm$ & 1.1 & 8.2 & $\pm$ & 0.7\\
geodesic loss & 9.2 & $\pm$ & 1.5 & 19.0 & $\pm$ & 2.3 & 20.7 & $\pm$ & 5.3 & 7.2 & $\pm$ & 0.4 & 8.9 & $\pm$ & 0.6 & 18.7 & $\pm$ & 3.2 & 11.0 & $\pm$ & 1.9 & 12.4 & $\pm$ & 3.4 & 10.2 & $\pm$ & 2.4\\
no artifact augm. & 17.4 & $\pm$ & 2.3 & 20.2 & $\pm$ & 2.0 & 18.3 & $\pm$ & 3.6 & 12.5 & $\pm$ & 2.5 & 10.5 & $\pm$ & 1.1 & 26.5 & $\pm$ & 3.8 & 11.3 & $\pm$ & 1.8 & 9.7 & $\pm$ & 1.8 & 13.6 & $\pm$ & 2.0\\
\hline
\end{tabular}}
\label{tab:navigators_ablations_aad}
\end{table*}

In Tables~\ref{tab:baselines}-~\ref{tab:translation}, we test for significance at $p\!<0.05$ using the Bonferroni-corrected pairwise Wilcoxon test for Research-Fetal, dHCP, and Clinical-Young test datasets. For Navigators, we use hierarchical permutation testing to account for time-series- and subject-level correlations.

To provide further insight, we also report performance statistics in Navigators aggregated by subject. Tables~\ref{tab:navigators_baselines_rot_error}-\ref{tab:navigators_baselines_trans_error} show that E(3)-Pose consistently outperforms the baselines across subjects and training datasets. Tables~\ref{tab:navigators_ablations_rot_error} and~\ref{tab:navigators_ablations_aad} show that regardless of the training data, E(3)-Pose ranks first, compared to all ablations, in the greatest number of subjects. Importantly, we highlight that for several subjects (e.g., S2 and S6), E(3)-Pose outperforms all ablations, which report larger errors in pose estimation. From visual inspection, we observe that these cases exhibit higher inherent pose ambiguity.

\subsubsection{Additional Examples.} 
\begin{figure*}
\centering
\centerline{\includegraphics[width=\textwidth]{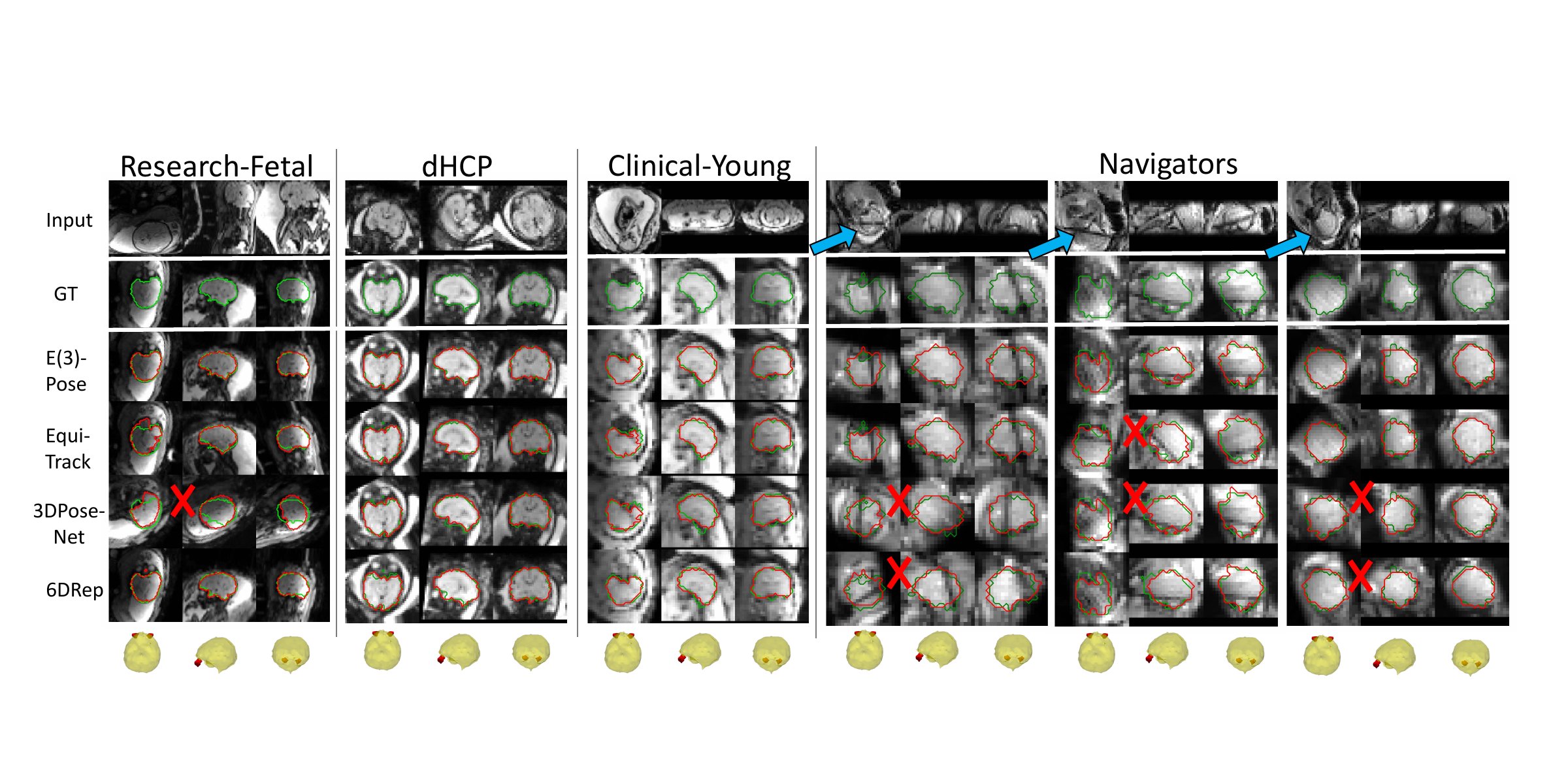}}
\caption{\textbf{Example results comparing E(3)-Pose to EquiTrack, 3DPose-Net, and 6DRep, trained on Research-Fetal.} See Fig.~\ref{fig:examples} in the main paper for descriptions.}
\label{fig:examples_methods}
\end{figure*}

\begin{figure*}
\centering
\centerline{\includegraphics[width=\textwidth]{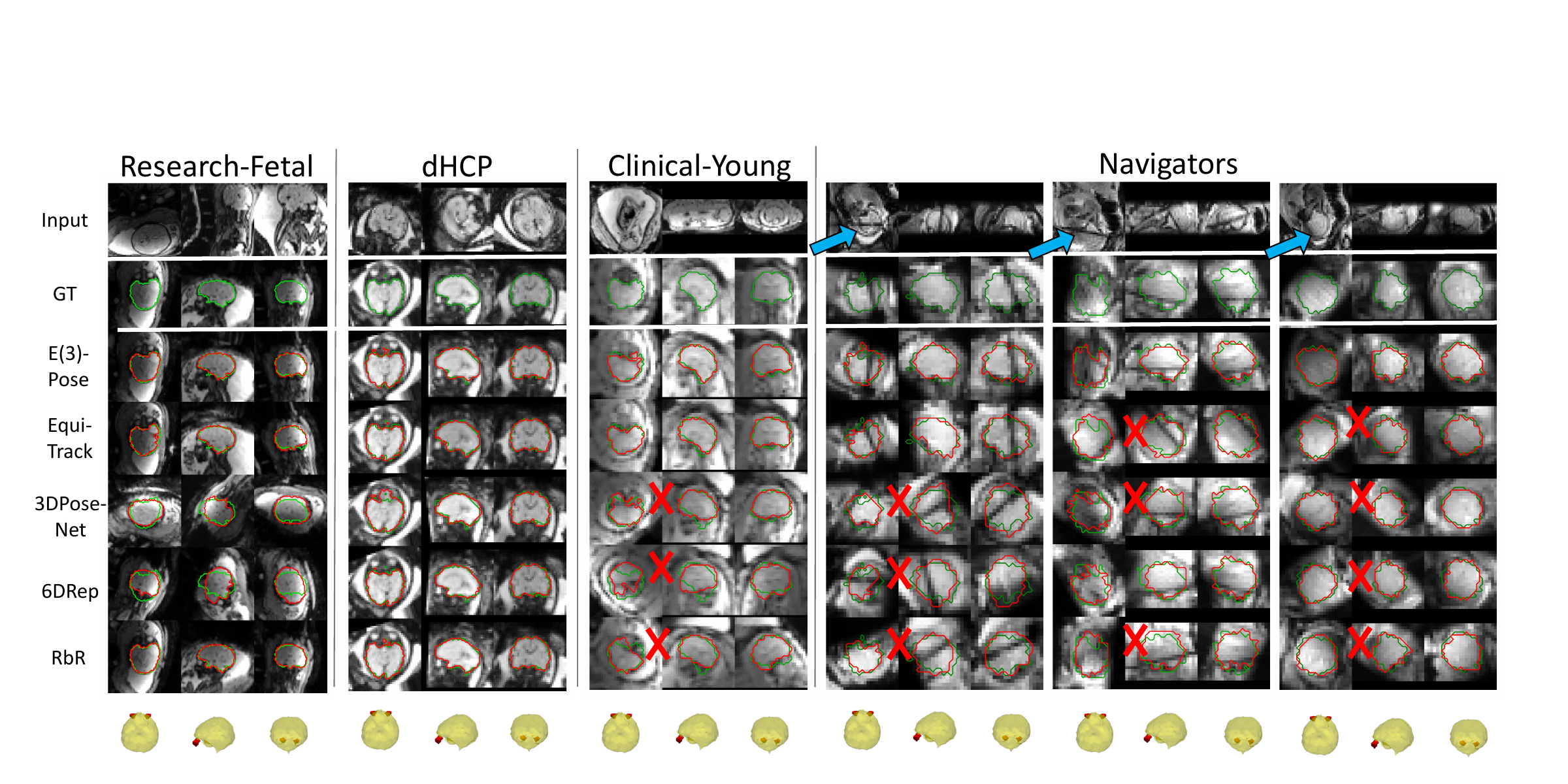}}
\caption{\textbf{Example results for methods trained on dHCP.} We omit FireANTs and Fetal-Align since these methods are not trained on dHCP. See Fig.~\ref{fig:examples} in the main paper for descriptions.}
\label{fig:examples_dhcp}
\end{figure*}

\begin{figure*}
\centering
\centerline{\includegraphics[width=\textwidth]{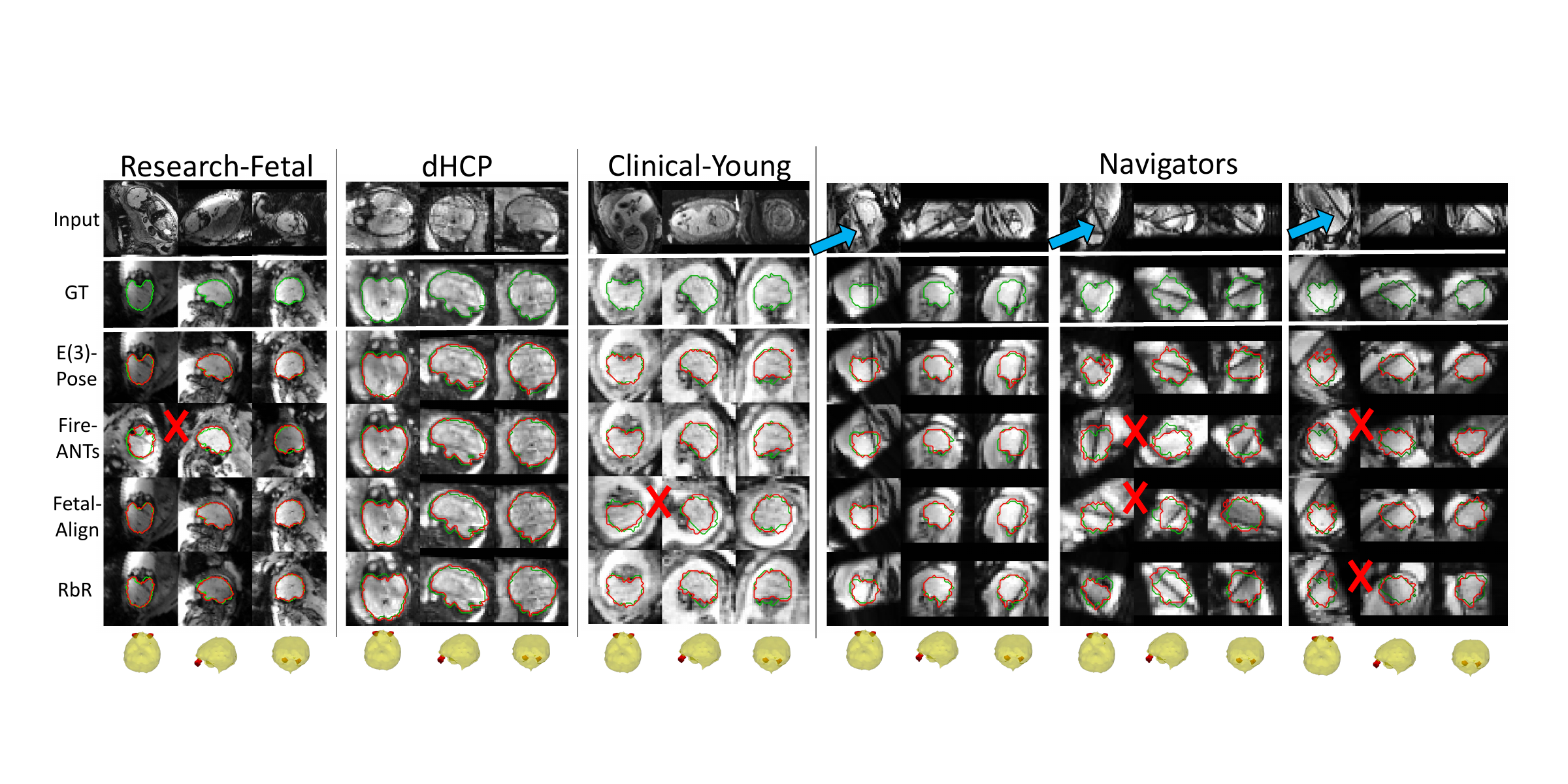}}
\caption{\textbf{Additional example results for methods trained on Research-Fetal.} See Fig.~\ref{fig:examples} in the main paper for descriptions.}
\label{fig:examples_more}
\end{figure*}

\begin{figure*}
\centering
\centerline{\includegraphics[width=\textwidth]{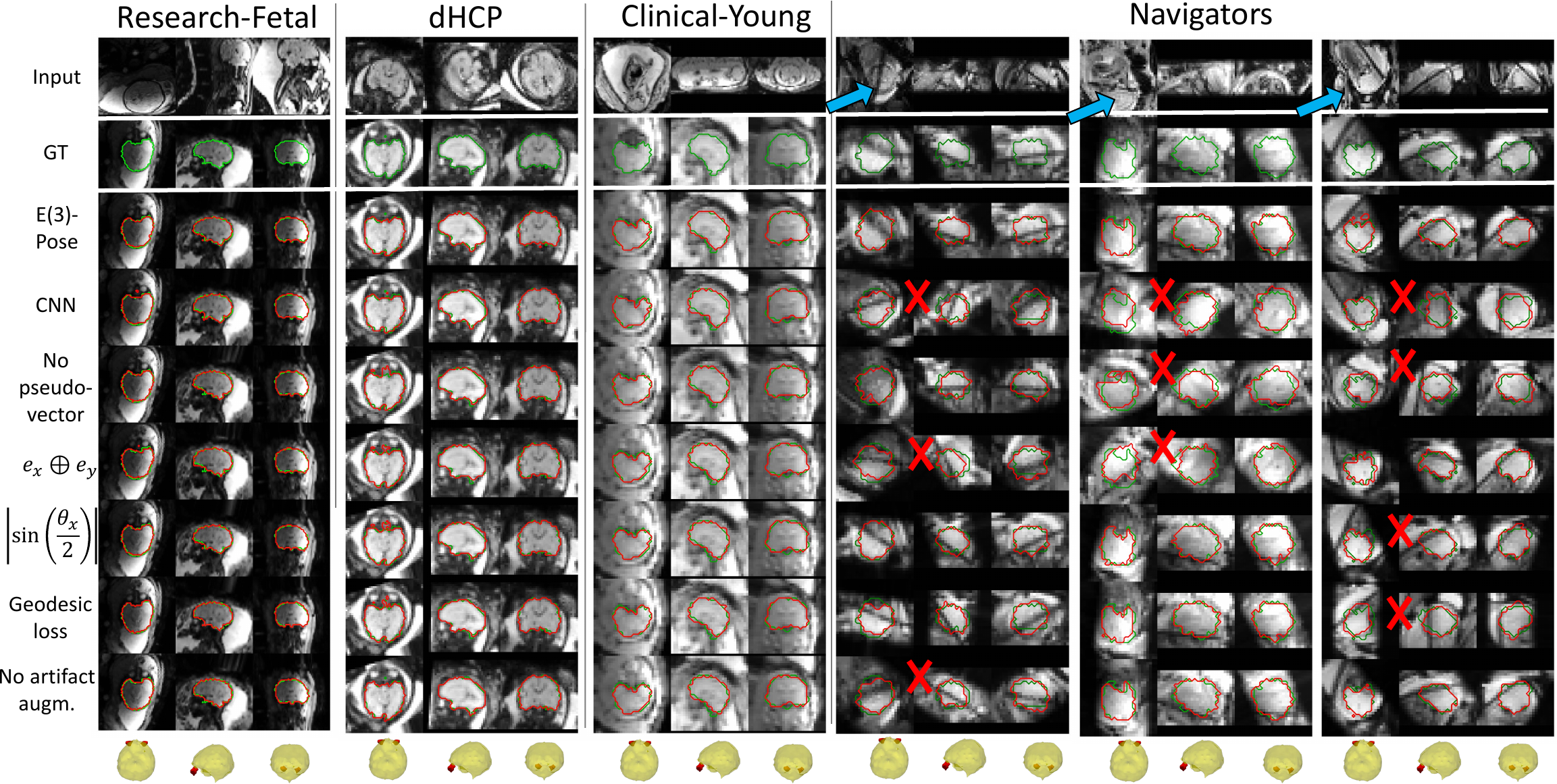}}
\caption{\textbf{Example results for ablations trained on Research-Fetal.} See Fig.~\ref{fig:examples} in the main paper for descriptions.}
\label{fig:examples_ablations}
\end{figure*}

In this section, we provide additional example alignments. Fig.~\ref{fig:examples_methods} displays the same volumes aligned by the baseline methods that were omitted from Fig.~\ref{fig:examples} in the main paper, i.e., EquiTrack, 3DPose-Net, and 6DRep. Fig.~\ref{fig:examples_dhcp} displays the same set of volumes aligned by all methods trained on dHCP. Fig.~\ref{fig:examples_more} replicates Fig.~\ref{fig:examples} for a different set of test volumes in each dataset. Fig.~\ref{fig:examples_ablations} provides example alignments for E(3)-Pose and all ablations.

\section{Ablation Study}
\label{sec:ablation_supp}

\subsubsection{Alternative Loss Functions.} In this section, we show why the first selected loss function for our ablation experiments does not respect the pseudovector symmetry of $\hat{e}_x$.

\begin{theorem}
Let $\mathcal{L'}((e_x,e_y,e_z),(\hat{e}_x,\hat{e}_y,\hat{e}_z))=\sum_{k\in\{x,y,z\}}|\sin{\frac{\theta_k}{2}}|$ be the modified training objective in the ablation study where we replace the term $|\sin{\theta_x}|$ with $|\sin{\frac{\theta_x}{2}}|$. Then, $\mathcal{L'}((e_x,e_y,e_z),(\hat{e}_x,\hat{e}_y,\hat{e}_z))\neq \mathcal{L'}((e_x,e_y,e_z),(-\hat{e}_x,\hat{e}_y,\hat{e}_z))$.
\end{theorem}
\begin{proof}
\begin{align*}
\mathcal{L'}((e_x,e_y,e_z),(\hat{e}_x,\hat{e}_y,\hat{e}_z)) &=|\sin{(\theta_x/2)}| + |\sin{(\theta_y/2)}| + |\sin{(\theta_z/2})|\\
&=\sqrt{\frac{1-\hat{e}_x\cdot e_x}{2}} + \sqrt{\frac{1-\hat{e}_y\cdot e_y}{2}} +\sqrt{\frac{1-\hat{e}_z\cdot e_z}{2}}\\
\mathcal{L'}((e_x,e_y,e_z),(-\hat{e}_x,\hat{e}_y,\hat{e}_z)) &=\sqrt{\frac{1+\hat{e}_x\cdot e_x}{2}} + \sqrt{\frac{1-\hat{e}_y\cdot e_y}{2}} +\sqrt{\frac{1-\hat{e}_z\cdot e_z}{2}}
\end{align*}
$\mathcal{L'}((e_x,e_y,e_z),(\hat{e}_x,\hat{e}_y,\hat{e}_z))= \mathcal{L'}((e_x,e_y,e_z),(-\hat{e}_x,\hat{e}_y,\hat{e}_z)) \iff \hat{e}_x\cdot e_x\!=\!0$.\\
Since $\hat{e}_x\cdot e_x\!=\!0$ is not true for any pair of unit vectors, it does not hold that $\mathcal{L'}((e_x,e_y,e_z),(\hat{e}_x,\hat{e}_y,\hat{e}_z))= \mathcal{L'}((e_x,e_y,e_z),(-\hat{e}_x,\hat{e}_y,\hat{e}_z))$.
\end{proof}
While the geodesic loss function does respect the pseudovector symmetry of $\hat{e}_x$, we found that training with this objective diminished performance on our data. The reasons for this are uncertain, but prior work suggests that the conversion of the network output to a proper rotation via SVD during training can cause gradient instabilities when the singular values are degenerate or close to 0~\cite{levinson2020}.

\subsubsection{Results on Research Data.} Table~\ref{tab:ablation_research} reports the performance statistics on research data for the ablations corresponding to Table~\ref{tab:ablation} in the main paper. For most experiments, there is no significant difference in performance metrics between E(3)-Pose and ablations. These results suggest that modeling inherent symmetries by construction yields competitive accuracy but does not improve performance on in-distribution, high-SNR volumes. Specifically, methods that do not explicitly address pose ambiguities perform well when such ambiguities are eliminated by clearly visible anatomy. The design choices in E(3)-Pose provide the largest performance gains in cases with low visibility and high uncertainty, which is more common in clinical data.

\begin{table*}
\centering
\setlength\tabcolsep{0pt}
\fontsize{7}{8.5}\selectfont
\caption{\textbf{Ablation performance statistics on research data.} Mean $\pm$ standard deviation statistics for rotation error ($^\circ$) and average absolute error (AAD, mm) are reported. Best score is in bold. * indicates statistical significance compared to E(3)-Pose ($p\!<\!0.05$, pairwise Wilcoxon test).}
\resizebox{\textwidth}{!}{%
\begin{tabular}{@{}l@{\;}
  r@{\,\(\pm\)\,}l@{\;}r@{\,\(\pm\)\,}l
  r@{\,\(\pm\)\,}l@{\;}r@{\,\(\pm\)\,}l
  r@{\,\(\pm\)\,}l@{\;}r@{\,\(\pm\)\,}l
  r@{\,\(\pm\)\,}l@{\;}r@{\,\(\pm\)\,}l@{}}
\toprule
& \multicolumn{8}{c}{\textbf{Trained on Research-Fetal}}
& \multicolumn{8}{c}{\textbf{Trained on dHCP}} \\
\cmidrule(lr){2-9}\cmidrule(lr){10-17}
& \multicolumn{4}{c}{\textbf{Research-Fetal test}} & \multicolumn{4}{c}{\textbf{dHCP}}
& \multicolumn{4}{c}{\textbf{Research-Fetal test}} & \multicolumn{4}{c}{\textbf{dHCP}} \\
\cmidrule(lr){2-5}\cmidrule(lr){6-9}\cmidrule(lr){10-13}\cmidrule(lr){14-17}
& \multicolumn{2}{c}{Rot. err.} & \multicolumn{2}{c}{AAD}
& \multicolumn{2}{c}{Rot. err.} & \multicolumn{2}{c}{AAD}
& \multicolumn{2}{c}{Rot. err.} & \multicolumn{2}{c}{AAD}
& \multicolumn{2}{c}{Rot. err.} & \multicolumn{2}{c}{AAD} \\
\midrule
E(3)-Pose (ours) & $5.1$ & $2.6$ & $3.0$ & $1.7$ & $7.4$ & $3.6$ & $3.7$ & 1.8 & $\mathbf{5.7}$ & $\mathbf{3.6}$ & $\mathbf{3.4}$ & $\mathbf{2.0}$ & $7.3$ & $3.4$ & $3.7$ & $1.7$ \\
\hline 
Standard CNN & $6.9$ & $3.5^*$ & $3.8$ & $1.9^*$ & $7.5$ & $4.2$ & $3.7$ & $2.0$ & $11.2$ & $4.9^*$ & $6.2$ & $3.1^*$ & $7.5$ & $3.4$ & $3.7$ & $1.7$\\
\hline 
No pseudovector  & $5.6$ & $2.4$ & $3.2$ & $1.6$ & $8.6$ & $3.3$ & $4.2$ & $1.7$ & $5.9$ & $3.0$ & $3.4$ & $1.8$ & $\mathbf{6.8}$ & $\mathbf{3.2}$ & $\mathbf{3.4}$ & $\mathbf{1.6^*}$\\
$h(R)\!=\!e_y\!\oplus\!e_z$ & $5.3$ & $2.5$ & $3.1$ & $1.5$ & $6.8$ & $3.5$ & $3.4$ & $1.6$ & $5.9$ & $2.7$ & $3.4$ & $1.6$ & $7.5$ & $3.4$ & $3.7$ & $1.7$\\
\hline 
$|\sin{\frac{\theta_x}{2}}|$ & $\mathbf{4.9}$ & $\mathbf{2.4}$ & $\mathbf{2.9}$ & $\mathbf{1.5}$ & $7.2$ & $3.6$ & $3.5$ & $1.6$ & $5.9$ & $3.2$ & $3.4$ & $1.8$ & $7.3$ & $3.2$ & $3.6$ & $1.5$\\
Geodesic loss & $\mathbf{4.9}$ & $\mathbf{2.1}$ & $\mathbf{2.9}$ & $\mathbf{1.2}$ & $\mathbf{6.5}$ & $\mathbf{3.8^*}$ & $3.3$ & $1.7^*$ & $14.7$ & $38.1$ & $6.8$ & $14.9$ & $73.8$ & $84.4^*$ & $24.9$ & $27.4^*$\\
No artifact augm. & $5.1$ & $2.2$ & $3.0$ & $1.4$ & $6.6$ & $3.2^*$ & $\mathbf{3.2}$ & $\mathbf{1.5^*}$ & $7.2$ & $3.9^*$ & $4.1$ & $2.4^*$ & $7.6$ & $3.4$ & $3.8$ & $1.6$\\
\hline
\end{tabular}}
\label{tab:ablation_research}
\end{table*}

\subsubsection{Analysis of Training Efficiency.}

\begin{figure*}
\centering
\centerline{\includegraphics[width=0.5\textwidth]{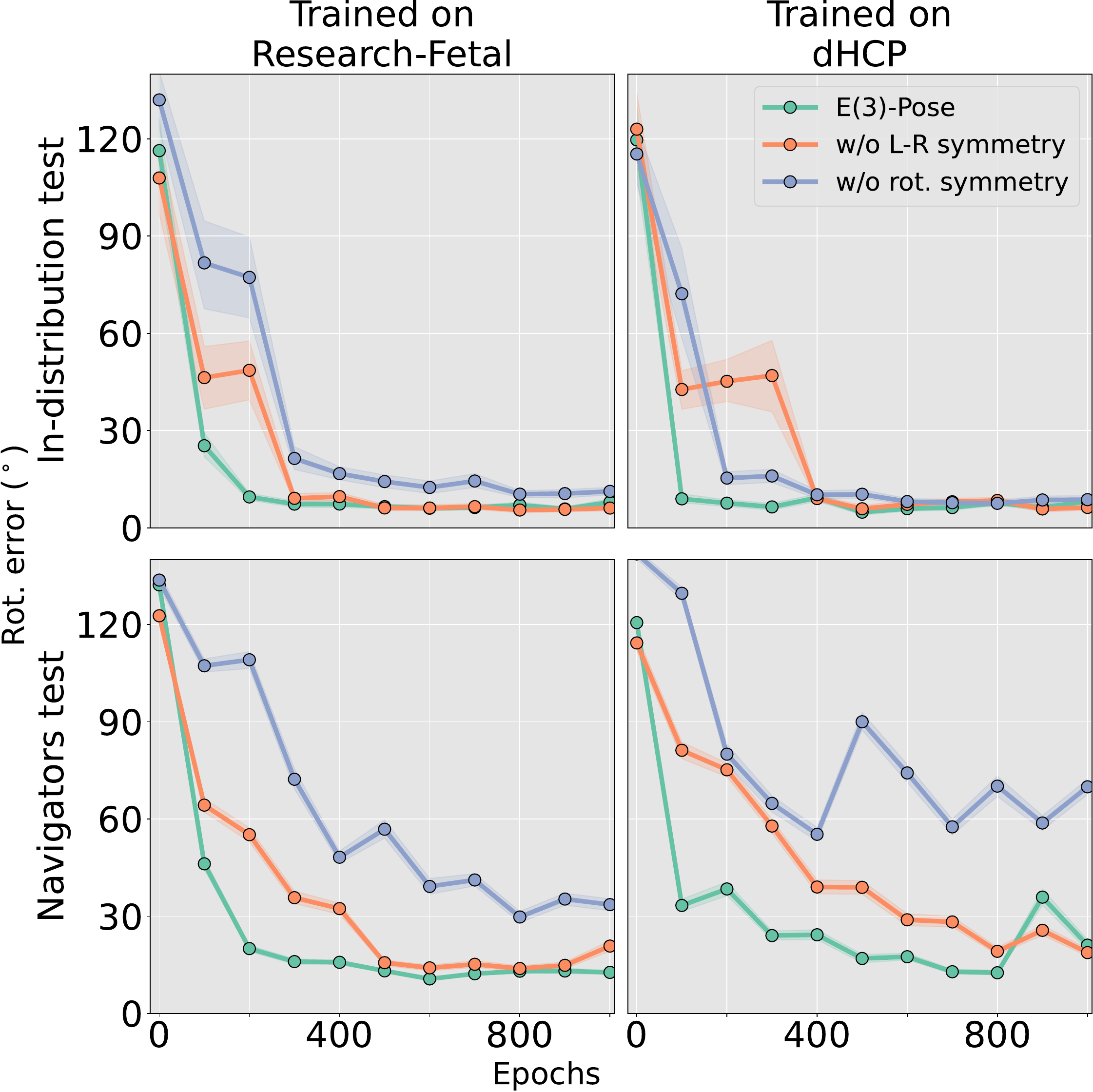}}
\caption{\textbf{Analysis of training efficiency.} Rotation error ($^\circ$) statistics on in-distribution (\textit{top}) and out-of-distribution ( \textit{bottom}, Navigators) test data as a function of the first 1000 epochs of training, for instances trained on Research-Fetal (\textit{left}) and dHCP (\textit{right}) training datasets. Compared to the ablations that do not model left-right head symmetry with a pseudovector and pose equivariance with rotation-equivariant convolutions, the performance of E(3)-Pose displays the fastest convergence on both in-distribution and out-of-distribution test datasets, and across training settings.}
\label{fig:convergence}
\end{figure*}

We now examine the impact of inductive biases in the network architecture on the training efficiency. Specifically, we study performance as a function of training time and compare our method to the variants of E(3)-Pose that do not model left-right head symmetry and pose equivariance under rotations, i.e., the pseudovector and standard CNN ablations, respectively. Tables~\ref{tab:ablation} and~\ref{tab:ablation_research} show that under sufficient training time, E(3)-Pose outperforms both ablations only on out-of-distribution data, with limited benefits on in-distribution test data. However, Fig.~\ref{fig:convergence} shows that the performance of E(3)-Pose also converges faster on \textit{both} in-distribution and out-of-distribution test data, across training environments. Our results suggest that restricting the hypothesis space based on inductive biases of the problem provides advantages in both training efficiency and generalizability~\cite{baxter2000}.

\section{Simulation Study}
\label{sec:sim}

\subsubsection{Additional Implementation Details.} We simulate stacks of 2D diagnostic slices (1$\!\times \!1$mm pixels, 3mm slice thickness) in a specific target anatomical orientation (i.e., left$\rightarrow$right, posterior$\rightarrow$anterior, or inferior$\rightarrow$superior). We simulate navigator volumes rapidly acquired before each slice from test volumes in Research-Fetal, by augmenting with 6mm isotropic resolution and a simulated spin history artifact based on the imaging plane parameters of the preceding slice. We simulate one stack in each orientation for each test volume, resulting in 168 total stacks. The number of slices per stack is adjusted for brain size as in current clinical practice, resulting in 23-40 slices per stack.

Recall from Appendix~\ref{sec:interleaved} that every slice is prescribed according to $\tilde{P}_k\!=\!\hat{T}_kP_k$, where $\hat{T}_k$ is the estimated head pose in the preceding navigator volume. We compare $\hat{T}_k$ returned by E(3)-Pose to that of ``motion-blind" slice prescription, where we set $\hat{T}_k\!=\!T_{k=0}$ for all $k\!\geq\!0$. To simulate inter-slice fetal brain motion for $k\!>\!0$, we sample rigid motion $T_{k+1}T_k^{-1}$ from real fetal head motion trajectories, interpolated at time intervals of TR=3s to match the time between navigator volume acquisitions in the clinical setting~\cite{shoemake1985}. Our motion trajectory dataset consists of full time-series from subjects in Clinical-Young, since increased uterine mobility in younger fetuses produces wider ranges of motion for our analysis. We follow prior work~\cite{xu2022} in annotating motion trajectories in this dataset for the purposes of simulation. Specifically, we apply an existing fetal landmark detection algorithm to detect the fetal eyes and shoulders in every frame~\cite{diaz2025}, manually correct algorithmically estimated landmarks, and finally compute the rigid motion between consecutive frames using detected landmarks.

We define the slice coverage distribution $p_C$ over the underlying brain volume on a 1mm resolution grid as the sum of the Gaussian kernels representing the point spread function~\cite{rousseau2006, kuklisovamurgasova2012, xu2022} of all slices in the stack. We then compute the coverage gap as the proportion of the brain volume where $p_C\!=\!0$.

\begin{figure*}
\centering
\centerline{\includegraphics[width=\textwidth]{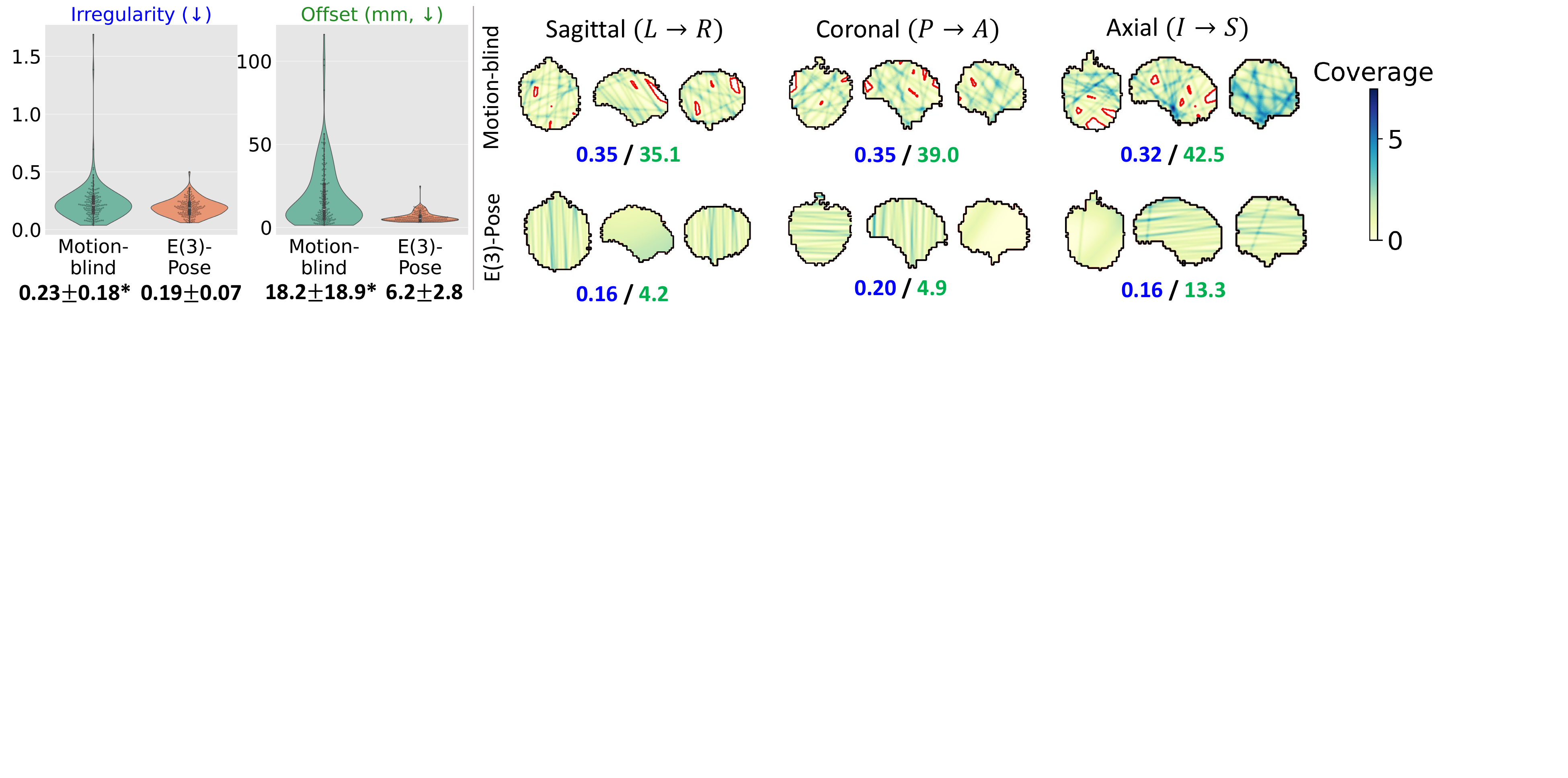}}
\caption{\textbf{Additional simulation results.} \textit{Left:} Coverage irregularity and slice offset (mm) of diagnostic slice stacks obtained using motion-blind prescription and E(3)-Pose. Mean $\pm$ standard deviation statistics are displayed. * indicates statistical significance ($p\!<\!0.05$, pairwise Wilcoxon). \textit{Right:} See Fig.~\ref{fig:sim} in the main paper for details. Here, coverage irregularity (blue) and slice offset (green, mm) metrics are respectively displayed.}
\label{fig:sim_supp}
\end{figure*}

\subsubsection{Additional Evaluations.} The primary aims of automated slice prescription are improvements in the spatial coverage of the brain and the radiological interpretability of the acquired stack of slices. In the main paper, we partially quantify these qualities with the coverage gap and slice obliqueness, respectively. In this section, we include two additional metrics that further assess the diagnostic potential of the simulated slice stacks. First, because uneven spatial coverage is correlated with poor coverage quality and coverage gaps, we calculate the coverage irregularity $\text{KL}(p_C||p_U)$, where KL is the Kullback-Leibler divergence and $p_U$ is the uniform distribution over the underlying brain volume. Second, since slice obliqueness only quantifies the rotational incoherence of slices within each stack, we also assess translational incoherence by computing the slice offset, i.e., the distance in mm between the center voxels of the GT and prescribed slices. Fig.~\ref{fig:sim_supp} demonstrates that in our simulations, slice prescription with E(3)-Pose significantly improves both additional quantitative metrics of diagnostic potential, compared to motion-blind prescription. 

\section{Additional Limitations and Future Work}
\label{sec:future_work}

\subsubsection{Navigators Dataset.} Although we evaluate E(3)-Pose on data representative of automated slice prescription with respect to acquisition parameters and image quality, our analysis is limited to 9 pregnant participants who were recruited at the same institution and imaged on the same scanner. Furthermore, Navigators does not include sufficient representation of second-trimester fetuses. To investigate the broader applicability of our method for clinical translation, future work will extend our analysis on navigator volumes acquired in a larger, more diverse cohort of pregnant volunteers. 

\subsubsection{Symmetries.} Beyond fetal imaging, it is interesting to investigate applications of our method under more relaxed reflectional and/or rigid symmetries~\cite{wang2022}, such as cardiac, lung, and adult brain imaging. Lastly, we will investigate whether our method can be adapted to objects with more complex (e.g., $n$-fold rotations) symmetries (Appendix~\ref{sec:rotational_symmetries}).

\section{Rotational Object Symmetries}
\label{sec:rotational_symmetries}

In this work, we have demonstrated a rotation-equivariant framework for pose estimation of objects with reflectional symmetries, e.g., the fetal head. However, many real-world objects possess more complex symmetries. In this section, we sketch a method to estimate the pose of objects with $N$-fold rotational symmetries about an axis with an E(3)-CNN. In future work, we will investigate whether the proposed method in future work stabilized pose estimation in high-ambiguity situations with rotational object symmetries.

\subsubsection{Rotation Parametrization.} Recall that we denote the rotation of the object frame relative to the input volume as $R$, the rotation parametrization as $h(R)$, and object symmetry group as $G_{\text{symm}}$. Here we consider the case where $G_{\text{symm}}\!=\!\{g_{\theta=2\pi(k/N)}| k=1,...,N\}$ where $\theta=2\pi(k/N)$ are the $N$-fold rotations in the plane of rotational symmetry. Let $e_\circlearrowleft$ be axis of rotational symmetry, and let parity($N$) be equal to the parity of the integer $N$. Then, we can train an E(3)-CNN to predict the following rotation parametrization:
\begin{equation}
\begin{aligned}
h(R)&=e_x\oplus e_y\oplus e_z,\\
&\text{s.t. }  e_x\!=\!D^{l=N}_{:,-N}(R),e_y\!=\!D^{l=N}_{:,N}(R) \hspace{0.2cm} \text{and} \hspace{0.2cm} e_z\!=\!e_{\circlearrowleft },\\
\rho_{h}(g_r)\!&=D^{l=N}(g_r)\oplus D^{l=N}(g_r)\oplus M(g_r)\\
&=\!\rho^{l=N}_{\text{parity}(N)}(g_r)\oplus \rho^{l=N}_{\text{parity}(N)}(g_r) \oplus \rho^{l=1}_{\odd}(g_r),
\end{aligned}
\label{eq:rot_param2}
\end{equation}
where $D^{l}:\text{SO(3)}\rightarrow \mathbb{R}^{2l+1\times2l+1}$ is the Wigner-D matrix function of order $l$, and $D^{l}_{:,m}$ denotes its columns, indexed from $-l\!\leq\!m\!\leq\!l$. Columns of the Wigner-D matrices are irreducible tensors with irreducible Wigner-D matrix representations~\cite{weiler2018,geiger2022}. Since $h(R)$ is parametrized in terms of irreducible tensors, it maintains equivariance under E(3). Therefore, the E(3)-CNN output is formulated as 1 vector and 2 irreducible tensors of order $l\!=\!N$ and parity $p\!=\!\text{parity}(N)$.

\begin{lemma}
\label{lemma:rot_symm}
\begin{align*}
\sum\limits_{-N\leq m' \leq N}[e_y]_{m'} Y^{l=N}_{m=m'}(\theta,\phi)\!=\!\sum\limits_{-N\leq m' \leq N}[D^{l=N}(g_{\theta\!=\!2\pi(k/N)})e_y]_{m'}Y^{l=N}_{m=m'}(\theta,\phi),
\end{align*}
where $Y^l_m$ is the spherical harmonic function of degree $l$ and order $m$, and the spherical coordinates $(\theta,\phi)$ refer to the polar and azimuthal angles, respectively. The same reasoning can be applied to $e_x$.
\end{lemma}

\begin{proof}
    Let $(\theta,\phi)\xrightarrow{R}(\theta',\phi')$ define the spherical coordinate transformation under rotation $R\in\text{SO(3)}$. We observe that
\begin{align*}
    \sum\limits_{-N\leq m' \leq N}[e_y]_{m'} Y^{l=N}_{m=m'}(\theta,\phi)
    &=\!\sum\limits_{-N\leq m' \leq N}D^{l=N}_{m',N}(R) Y^{l=N}_{m=m'}(\theta,\phi)=Y^{l=N}_{m=N}(\theta',\phi'),
\end{align*}  
where we use the property $Y_m(\theta ',\phi ')\!=\!\sum\limits_{-l\leq m'\leq l}D^l_{m'm}(R)Y^l_{m'}(\theta,\phi)$ to obtain the second equality. It follows that
\begin{align*}
\sum\limits_{-N\leq m' \leq N}[e_y]_{m'} Y^{l=N}_{m=m'}(\theta,\phi)&=Y^{l=N}_{m=N}(\theta',\phi')\\
&=\!e^{i2\pi k}Y^{l=N}_{m=N}(\theta',\phi') \text{   for any integer } k \\
&=\!e^{iN2\pi(k/N)}Y^{l=N}_{m=N}(\theta',\phi')\\
&=\!Y_m^{l=N}(\theta'+\frac{2\pi k}{N},\phi')
\end{align*}
where we use the property $Y_m^l(\theta_1\!+\!\theta_2,\phi)\!=\!e^{im\theta_2}Y_m^l(\theta_1,\phi)$ to obtain the last equality. Using this result and the previously stated property of spherical harmonic functions under rotation by Wigner-D matrices, we have
\begin{align*}
\sum\limits_{-N\leq m' \leq N}[e_y]_{m'}&Y^{l=N}_{m=m'}(\theta,\phi)\!=\!Y_m^{l=N}(\theta'+\frac{2\pi k}{N},\phi')\\
&=\sum\limits_{-N\leq m' \leq N}D^{l=N}_{m',N}(g_{\theta\!=\!2\pi(k/N)}R)Y^{l=N}_{m=m'}(\theta,\phi)\\
&=\sum\limits_{-N\leq m' \leq N}[D^{l=N}(g_{\theta\!=\!2\pi(k/N)})D^{l=N}_{:,N}(R)]_{m'}Y^{l=N}_{m=m'}(\theta,\phi)\\
    &=\sum\limits_{-N\leq m' \leq N}[D^{l=N}(g_{\theta\!=\!2\pi(k/N)})e_y]_{m'}Y^{l=N}_{m=m'}(\theta,\phi),
\end{align*}
where we use the property $D^l(R_1)D^l(R_2)\!=\!D^l(R_1R_2)$ for any $R_1,R_2\!\in\!$ SO(3) to obtain the third equality.
\end{proof}

\begin{theorem}
$h(R)\!=\!\rho_h(g_{\theta\!=\!2\pi(k/N)})h(R)$ for all $k\!=\!1,...,N$.
\end{theorem}

\begin{proof}
First, it is easy to see that $e_z\!=\!M(g_{\theta\!=\!2\pi(k/N)})e_z$ because in-plane rotations leave orthogonal vectors unchanged. Second, the result of Lemma~\ref{lemma:rot_symm} indicates that $e_x\!=\!D^{l=N}(g_{\theta\!=\!2\pi(k/N)})e_x$ and $e_y\!=\!D^{l=N}(g_{\theta\!=\!2\pi(k/N)})e_y$.
\end{proof}

\subsubsection{Training.} We aim to respect the symmetry of the rotation parametrization in the training objective. To this end, any standard regression loss (e.g., $\mathcal{L}_1$ or $\mathcal{L}_2$ norm) on the error term $h(R)-[\hat{e}_x\oplus \hat{e}_y\oplus \hat{e}_z]$ between the GT rotation parametrization and the E(3)-CNN output, respectively, satisfies this condition.

\subsubsection{Inference.} Since Wigner-D matrices are a Fourier basis for functions defined on SO(3), the E(3)-CNN output represents a likelihood function $P$ over SO(3), where $P(R')$ is monotonically related to the inner product between $h(R')$ and $[\hat{e}_x\oplus \hat{e}_y\oplus \hat{e}_z]$~\cite{lee2024}. It is possible to sample from $P$ by querying values on a predefined equivolumetric grid over SO(3)~\cite{gorski2005}. Indeed, if the E(3)-CNN output $[\hat{e}_x\oplus \hat{e}_y\oplus \hat{e}_z]$ is equal to $h(R)$, then $P$ is nonzero and uniformly distributed over the set of poses that are symmetrically equivalent to $R$, and zero elsewhere.
\bibliographystyle{splncs04}
\bibliography{main}
\end{document}